\documentclass[]{style}

\usepackage{amsmath}
\usepackage{amssymb}
\usepackage{mathtools}
\usepackage{multirow}
\usepackage{makecell}
\usepackage{tabularx}
\usepackage{enumitem}
\usepackage{wrapfig}
\usepackage{pifont}
\usepackage{url}

\definecolor{RowBlue}{HTML}{E9F2FB}
\definecolor{RowRed}{HTML}{F9EAEA}

\crefname{figure}{Fig.}{Figs.}
\renewcommand{\emph}[1]{\textit{#1}}

\usepackage{lmodern}\renewcommand{\sfdefault}{sans}

\usepackage{amsmath,amsfonts,bm}

\def\eqref#1{equation~\ref{#1}}

\def\1{\bm{1}}

\DeclareMathAlphabet{\mathsfit}{\encodingdefault}{\sfdefault}{m}{sl}
\SetMathAlphabet{\mathsfit}{bold}{\encodingdefault}{\sfdefault}{bx}{n}

\usepackage{booktabs}
\usepackage{array}
\usepackage{xspace}
\usepackage{adjustbox}
\usepackage{multirow}
\newcolumntype{L}[1]{>{\raggedright\arraybackslash}p{#1}}

\newif\ifwidelayout
\usepackage{newfloat}
\DeclareFloatingEnvironment[name=Algorithm,placement=t]{algorithm}
\newcommand{\algcmt}[1]{\quad{\color{black!55}$\triangleright$\,#1}}
\newcommand{\algkw}[1]{\textbf{#1}}
\newcommand{\runin}[1]{{\bfseries #1}}  %
\usepackage{tcolorbox}
\newtcolorbox{algobox}{colback=black!2,colframe=black!30,boxrule=0.5pt,arc=2pt,
  left=4pt,right=4pt,top=1pt,bottom=1pt,boxsep=1pt}
\newcommand{\farmalgorithmwrap}{}
\newcommand{\farmalgorithmpre}{}
\newcommand{\farmalgobody}{\begin{algobox}\small
\begin{tabbing}
\hspace{1.6em}\=\hspace{1.1em}\=\hspace{1.1em}\=\kill
\algkw{Input:} episode $e^c$, request $(\ell,n,\mu)$, capability layer $\mathcal{K}$\\
1:\> $G \gets \textsc{Ground}(e^c)$ \algcmt{cameras, scale, objects}\\
2:\> $S^\star \gets \textsc{Qualify}(\text{simulator}, G)$ \algcmt{skill probes}\\
3:\> $T, J \gets \textsc{Contract}(G,\ell)$ \algcmt{hidden judge}\\
4:\> $C \gets \textsc{Compile}(T)$ on development seeds\\
5:\> admit $C$ only if $J$ passes on holdout seeds\\
6:\> $R \gets \textsc{RenderContract}(G,\mu)$ \algcmt{real references}\\
7:\> $\mathcal{D}\gets\emptyset$; \ $\mathcal{L}\gets\emptyset$ \algcmt{dataset, attempt ledger}\\
8:\> \algkw{while} $|\mathcal{D}| < n$ \algkw{do}\\
9:\>\> $E \gets \textsc{Rollout}(C, z)$ on $S^\star$ \algcmt{$z\sim$ setup sampler}\\
10:\>\> \algkw{if} $J(E)$ fails \algkw{or} replay differs \algkw{then}\\
11:\>\>\> $\mathcal{L}\gets\mathcal{L}\cup\{E\}$; \algkw{continue}\\
12:\>\> $O \gets \textsc{Render}(E, R)$\\
13:\>\> \algkw{if} $\textsc{Verify}(O,E)$ \algkw{and} $\textsc{Review}(O,E)$ \algkw{then}\\
14:\>\>\> $\mathcal{D}\gets\mathcal{D}\cup\{(O,E)\}$\\
15:\>\> \algkw{else} $\mathcal{L}\gets\mathcal{L}\cup\{O\}$; reroute to the earliest stage\\
16:\> \algkw{return} $\mathcal{D}$, $\mathcal{L}$ \algcmt{failed attempts are kept}
\end{tabbing}
\end{algobox}}
\newcommand{\farmalgorithm}{\begin{algorithm}[t]\caption{\ours for one task.}\label{alg:farm}%
\vspace{-0.3em}\farmalgobody\vspace{-0.6em}\end{algorithm}}
\usepackage{pifont}
\newcommand{\auOwn}{\textcolor{green!50!black}{\ding{108}}}
\newcommand{\auProp}{\textcolor{blue!60!black}{\ding{109}}}
\newcommand{\auNo}{\textcolor{red!70!black}{\ding{55}}}
\newcommand{\auNone}{\textcolor{black!35}{---}}
\newcommand{\ours}{\textsc{DataFarm}\xspace}
\providecommand{\colfigwidth}{\linewidth}
\providecommand{\coltabscale}{1}
\newlength{\coltabwidth}
\AtBeginDocument{\setlength{\coltabwidth}{\coltabscale\columnwidth}}

\newcommand{\papertitle}{PyRUA-\ours: LLM Agent as a Cost-Efficient Teleoperator for Real-Ready Robot Data}

\titleformat*{\paragraph}{\bfseries}
\widelayouttrue
\usepackage{needspace}
\newcommand{\teasercaption}{\textbf{What the farm delivers.}
\textbf{(a)}~The same frame of a simulator rollout and of our H3 rendering for each of the five tasks; the
rendering is photographic and keeps the simulated arm pose, objects, and contacts (insets enlarge the dashed regions).
\textbf{(b)}~From a request and one calibration episode, the agent builds the task in simulation, and the video
model renders each rollout in the original scene or in a new scene described by text.
\textbf{(c)}~Real-robot successes of policies trained only on farm data or on the same number of teleoperated
episodes, in the original scene and under unseen appearance.
\textbf{(d)}~Mean costs over the five tasks (Section~\ref{sec:res-cost}).}

\renewcommand{\colfigwidth}{0.58\linewidth}
\renewcommand{\coltabscale}{0.58}
\renewcommand{\farmalgorithmwrap}{\par\begin{wrapfigure}{r}{0.55\textwidth}\vspace{-1.2em}%
\captionof{algorithm}{\ours for one task.}\label{alg:farm}\vspace{-0.7em}\farmalgobody\vspace{-1.4em}\end{wrapfigure}}
\let\farmalgorithm\farmalgorithmwrap
\renewcommand{\farmalgorithmwrap}{}

\secondleftlogo{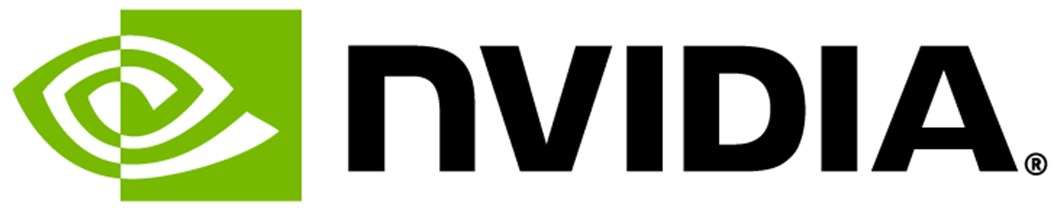}

\title{\papertitle}

\renewcommand{\authorfont}{\fontsize{11}{12.5}\selectfont}
\author[1,*]{Kewei Zhang}
\author[1,*]{Juncheng Ma}
\author[1]{Guodong Lv}
\author[2,\dagger]{Duomin Wang}
\author[1]{Feixiang Ruan}

\authorbreak
\author[1]{Ruiyang Si}
\author[2]{Chuanrui Zhang}

\author[1]{Yuxuan Chen}
\author[1]{Hao Chen}

\author[1]{Yichen Liu}

\authorbreak
\author[1]{Yintianrun Zhang}
\author[1]{Yufan Deng}
\author[2,\dagger]{Enze Xie}
\author[1,\dagger]{Daquan Zhou}
\affiliation[1]{Peking University}
\affiliation[2]{NVIDIA}
\contribution[*]{Equal contribution}
\contribution[\dagger]{Project lead}

\usepackage{fontawesome5}
\newcommand{\projectlink}[3]{\makebox[1.6em][l]{\normalsize #1}{\sffamily\bfseries\small #2:}~{\small\url{#3}}}
\abstract{Teleoperation remains the default way to collect robot demonstrations, but every
episode costs robot and operator time, and every new task starts from scratch. We ask
whether a coding agent can take over the teleoperator's job. In our \ours,
the agent receives a data request and one calibration episode of the robot in the target
scene and rebuilds the task in simulation: it writes the scene, an expert controller, and
a success judge. The simulator supplies exact actions, and a video model conditioned on
real images renders each rollout as photographic video, generating all camera views
together consistently. The output is real-ready data:
synchronized camera views labelled with the robot's actions, on
which a policy can be trained directly and deployed on the real robot. Independent judges
certify the expert and every rendered episode before admission. On five real-robot tasks, policies trained only
on farm data match policies trained on as many teleoperated episodes (93 vs.\ 84
successes in 100 trials), and under an unseen tablecloth and coloured light they keep
87 successes where the teleoperation-trained policies fall to 46. Adapting a task is
fully automatic and costs \$21 of API usage on average, paid once per task, plus \$5 for a
render contract; every further episode costs \$0.80 on average for simulation, rendering, and
review, with no robot or operator.
One hour of three-view data for a task costs \$107 on average, these costs included,
slightly below a quote of \$117 for teleoperation in China and below \$290
in labour alone in North America.
\par\vskip 3mm
\projectlink{\textcolor[HTML]{3A8DC5}{\faGlobeAmericas}}{Project Page}{https://dagroup-pku.github.io/PyRUA-Lean/DataFarm}\par\vskip 1.2mm
\projectlink{\faGithub}{GitHub Repo}{https://github.com/DAGroup-PKU/PyRUA-Lean/tree/DataFarm}\par\vskip 1.2mm
\makebox[1.6em][l]{\normalsize\faEnvelope[regular]}{\sffamily\bfseries\small Corresponding Author:}~{\small\href{mailto:zhoudaquan21@gmail.com}{\texttt{Daquan Zhou}}}}

\date{\today}

\begin{document}
\maketitle

\begin{figure}[!ht]
  \centering
  \includegraphics[width=\textwidth]{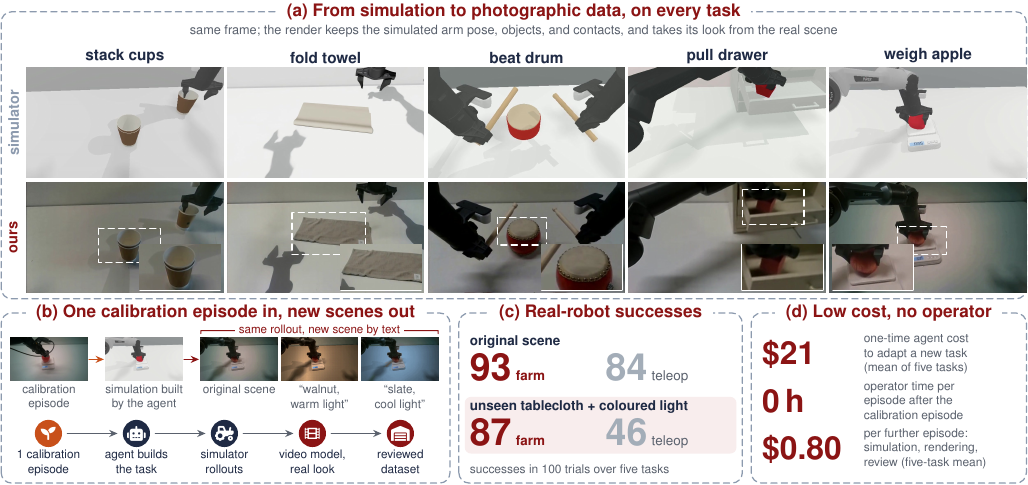}
  \caption[Overview]{\teasercaption}
  \vspace{-1em}
  \label{fig:overview}
\end{figure}

\section{Introduction}
\label{sec:introduction}

Teleoperation remains the primary source of training data for robot
policies~\citep{kim2024openvla,black2024pi0,black2025pi05}. Its appeal is that the
data are faithful by construction: every recorded action was executed by the robot,
and every image shows the scene in which the policy will be deployed. Its weakness
is cost. Each episode requires a robot, an operator, and resets after failures, so
the cost grows with the amount of data and starts over with every new task, scene,
or viewpoint.

Synthetic data promises to break this scaling, yet existing pipelines trade the cost
of collection for two other costs. The first is human engineering: whether
demonstrations are multiplied from a few seeds~\citep{mandlekar2023mimicgen,xue2025demogen}
or tasks are written by language models~\citep{wang2023gensim,chen2025robotwin2}, a
person still has to assemble and debug the scene, the expert, and the success check
for every new task. The second is the visual gap between simulation and reality.
Real-to-sim-to-real methods narrow it by reconstructing the
scene~\citep{torne2024rialto,han2025re3sim,qureshi2025splatsim}, but whatever is hard
to reconstruct, such as deformable cloth or a lit display, still looks synthetic,
while video generation models that synthesize the motion themselves can no longer
provide exact action labels~\citep{jang2025dreamgen}. Producing \emph{real-ready} robot data, exact enough to train on and realistic
enough to deploy on the real robot without any real demonstrations, cheaply for every
new task therefore remains an open problem.

We address this problem with \ours, an agentic data farm in which a coding agent
takes over the teleoperator's role, a simulator supplies the actions, and a video
model supplies the appearance (Figure~\ref{fig:pipeline}). Given a data request and a
single calibration episode, a short recording of the robot in the target scene, the
agent compiles the task into simulation as an engineer would: it reconstructs the
scene and cameras from the calibration episode, verifies that the simulator can
reproduce the required skills, and writes an expert controller together with a
success judge. The simulator then executes the expert from many initial states to
produce trajectories with exact actions and states, and a video model conditioned on
real frames renders each trajectory, all camera views in one generation, in the
appearance of the target scene, where new
appearances such as a different tablecloth or lighting are specified in text. The
output is real-ready: synchronized, mutually consistent photographic camera views with the robot's own
action labels, which a policy consumes exactly as it would consume teleoperated
data.
Because the agent harness is decoupled from individual tasks, a new task requires no
task-specific engineering by a person, and once a task is compiled, every further episode is
produced without a robot, an operator, or a reset.

Delegating data production to an agent, however, raises a question that
teleoperation never faces: who verifies the data? An agent that writes both the
expert and its success checks may approve its own mistakes, and a video model asked
to look realistic may copy the layout of its references, hallucinate objects, or
drift from the simulated motion. \ours therefore separates certification from
production. The expert is developed against a judge it cannot inspect and admitted
only on held-out initial states, the renderer may alter appearance but not geometry,
objects, or timing, and every rendered episode is compared with its simulated state
by machine checks and an independent reviewer before it enters the dataset.

Experiments on five bimanual tasks covering rigid, articulated, deformable, and
display-bearing objects show that this design yields real-ready data at low cost:
policies are trained on it directly and deployed on the real robot
(Figure~\ref{fig:overview}). Policies trained only on farm data
match those trained on the same number of teleoperated episodes in the original scene
(93 vs.\ 84 successes in 100 trials) and remain robust under a tablecloth and coloured
light absent from both training sets (87 vs.\ 46), which we attribute to the
appearance variation the farm adds around fixed, exact trajectories. Rendering with
real images is essential to this transfer: the same trajectories rendered with
conventional domain randomization yield only 26 and 21 successes
(Section~\ref{sec:res-policy}). Every camera view of a farm episode both follows the
simulation and looks real: image statistics place farm episodes more than twice as close
to real camera frames as the simulator's own renderings, and the farm's renderer keeps
all views faithful to the simulated motion, whereas calling the video model directly does
not (Section~\ref{sec:res-views}). The farm also operates without human
intervention: adapting a task costs \$21 of API usage on average, paid once per task, a render
contract adds \$5, and every further episode costs \$0.80 on average for simulation, rendering,
re-rendering, and review, in the
original scene or under new appearance, so 100 episodes of a task cost \$1.06 per
episode on average, and one hour of synchronized three-view data for a task costs \$107,
these costs included, slightly below a quote of \$117 for teleoperation in China and below
\$290 in labour alone in North America (Section~\ref{sec:res-cost}).

In summary, our contributions are threefold:
\begin{itemize}
  \item We introduce \ours, an agentic data farm that turns a data request and one
  calibration episode into verified, real-ready robot data without human intervention.
  The simulator determines actions and states, a real-image-conditioned video model
  determines appearance, and certification is kept separate from production, all under
  an agent harness that is independent of individual tasks.
  \item We show that the farm produces action-labelled, multi-view data that closely
  matches real camera recordings, at a lower cost than teleoperation.
  \item We show on a real robot that policies trained solely on farm data match
  teleoperated data in the original scene, generalize substantially better to unseen
  appearance, and far outperform a conventional real-to-sim-to-real baseline.
\end{itemize}

\section{Related Work}
\label{sec:related-work}

\paragraph{Collecting and multiplying demonstrations.}
Large teleoperated datasets drive recent vision-language-action
models~\citep{kim2024openvla,black2024pi0,black2025pi05}, and hand-held interfaces and
fleet orchestrators lower the cost of each episode~\citep{chi2024umi,ahn2024autort}, but
every episode still needs a person, a device, and a physical scene. Demonstration
generators multiply a few demonstrations into many, in
simulation~\citep{mandlekar2023mimicgen,garrett2024skillgen,jiang2025dexmimicgen} or in
point-cloud space~\citep{xue2025demogen}, assuming a simulation task that people have
built or point-cloud observations. \ours needs one calibration episode, reuses none of
its actions, and outputs photographic multi-view video.
\vspace{-1em}

\paragraph{Language models and agents that build tasks.}
Language models write policies as code~\citep{liang2023codeaspolicies},
rewards~\citep{ma2023eureka}, and simulation tasks with their
supervision~\citep{wang2023gensim,wang2023robogen,chen2025robotwin2}; coding agents
resolve real repository issues~\citep{yang2024sweagent} and improve their own
programs~\citep{zelikman2023stop,hu2024adas,zhang2025dgm}. These tasks are specified by
text and asset libraries rather than by a real scene, and the model that writes a task
often also writes its check, although language models are unreliable judges of their
own output~\citep{huang2024selfcorrect}. \ours grounds the task in a recording of the
real scene, has the agent write the expert against a judge it cannot see, and leaves
admission to independent reviewers.
\vspace{-0.5em}

\paragraph{Real-to-sim-to-real.}
Pipelines of this kind reconstruct the target scene as digital twins or
cousins~\citep{torne2024rialto,chen2024urdformer,dai2024digitalcousins,ranawaka2026simfoundry}
and make them photorealistic with Gaussian splatting or rendered
scans~\citep{qureshi2025splatsim,li2024robogsim,han2025re3sim,yu2025real2render2real,han2026video2sim2real}.
Their appearance is only as good as the reconstructed assets: every object must be
scanned or modelled, cloth and lit displays are hard to represent, and a new scene
needs a new capture. \ours keeps the simulator for actions and states but takes
appearance from a video model conditioned on real images.
\vspace{-0.5em}

\paragraph{Generative augmentation and video models.}
Image-level augmentation inpaints objects and backgrounds frame by
frame~\citep{yu2023rosie,chen2023genaug,yuan2025roboengine}. Video models generate robot
videos with inferred actions~\citep{jang2025dreamgen}, generate video under structural
control~\citep{nvidia2025cosmostransfer,liu2025robotransfer}, or turn simulator
renderings into realistic video~\citep{wang2025embodiedreamer,chen2026craft}, but
structure, physics, and task adherence can fail independently in such
videos~\citep{deng2026rbench}. \ours adds an agent that builds the simulation task from
one calibration episode, conditioning on that episode's own images, and independent
admission of every multi-view episode against its simulated state.
Table~\ref{tab:related} summarizes the comparison, and Appendix~\ref{sec:app-related} discusses each line of work in more detail.

\begin{table*}[t]
\centering
\footnotesize
\setlength{\tabcolsep}{3pt}
\resizebox{1.0\textwidth}{!}{%
\begin{tabular}{lllllc}
\toprule
Method & Input from the target task & Simulation task built by & Action labels & Observations & Real robot \\
\midrule
MimicGen~\citep{mandlekar2023mimicgen} & a few demos in simulation & people & simulator & simulator rendering & \checkmark \\
DemoGen~\citep{xue2025demogen} & one real demo & --- (no simulator) & edited demo & point clouds & \checkmark \\
RoboTwin~2.0~\citep{chen2025robotwin2} & task description, assets & language model & simulator & rendering + DR & \checkmark \\
Re$^3$Sim~\citep{han2025re3sim} & scene reconstruction & people & simulator & Gaussian splatting & \checkmark \\
SplatSim~\citep{qureshi2025splatsim} & object and scene scans & people & simulator & Gaussian splatting & \checkmark \\
Real2Render2Real~\citep{yu2025real2render2real} & object scans, human video & --- (no dynamics) & retargeted & rendered scans & \checkmark \\
DreamGen~\citep{jang2025dreamgen} & real teleoperated data & --- (no simulator) & inferred (IDM/LAM) & generated video & \checkmark \\
EmbodieDreamer~\citep{wang2025embodiedreamer} & real demos, real-to-sim calibration & people & replayed real actions & sim-conditioned video & \checkmark \\
CRAFT~\citep{chen2026craft} & a few real teleoperated demos & people (digital twin) & simulator & edge-conditioned video & \checkmark \\
\midrule
\ours (ours) & one calibration episode & agent & simulator & real-image-conditioned video & \checkmark \\
\bottomrule
\end{tabular}%
}
\caption{Comparison with related data-generation pipelines. Among them, only in
\ours does an agent build the simulation task from a single calibration episode of the
scene, reusing none of its actions, with observations rendered from that episode's own images
as appearance reference.}
\label{tab:related}
\end{table*}

\section{Problem Formulation}
\label{sec:problem}

\begin{figure*}[t]
  \centering
  \includegraphics[width=\textwidth]{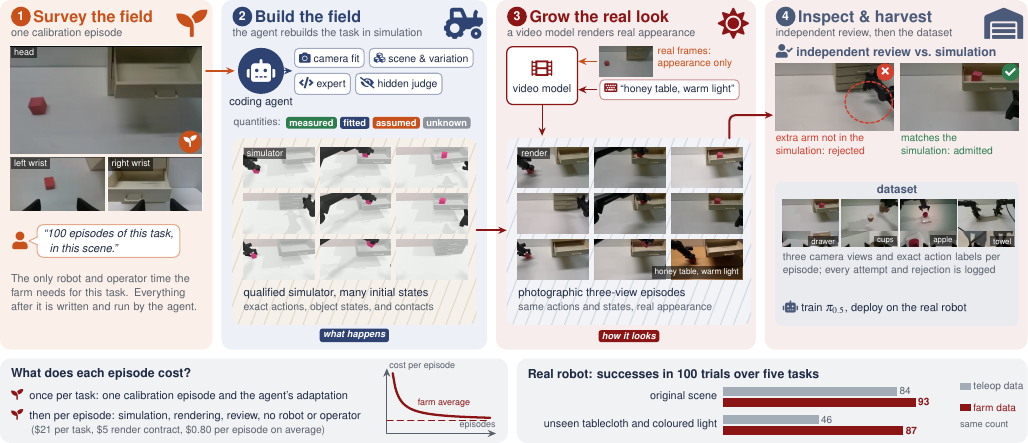}
  \caption{\textbf{Overview of \ours.}
\textbf{(1)}~A request and one calibration episode of the scene are the only inputs.
\textbf{(2)}~A coding agent rebuilds the task in simulation and writes an expert against a judge it cannot see; the simulator then runs the expert from many initial states and fixes the actions, object states, and contacts.
\textbf{(3)}~A video model conditioned on real frames renders every rollout in the scene's appearance, and a new appearance is a text change.
\textbf{(4)}~An independent reviewer compares each render with its simulated state, and only admitted episodes enter the dataset.
\textbf{Bottom:} the farm pays once per task rather than for every episode, and its data matches teleoperated data on the real robot. Shown on the pull-drawer task; the cost plot is schematic.}
  \label{fig:pipeline}
  \vspace{-1em}
\end{figure*}

\ours takes a data request and returns a dataset together with the evidence that
certifies it. A request $\rho=(e^c,\ell,n,\mu)$ consists of a calibration episode
$e^c$, a natural-language description $\ell$ of the task and its success condition,
the number of episodes $n$, and a data mode $\mu$. The calibration episode is a short
multi-view recording of the robot in the target scene, which identifies the scene and
the cameras but never serves as a source of actions. The data mode is either
$\textsc{sim2real}$, which reproduces the recorded scene, or
$\textsc{ood}(\mathcal{V})$, which additionally permits the appearance variations
$\mathcal{V}$, such as a new tablecloth. No adapter, judge, or prompt for the task
exists in advance. The farm delivers $n$ episodes $d_i=(o^i,a^i,x^i,k^i,y^i,q^i)$,
each consisting of photographic multi-view observations $o$, simulator actions $a$,
simulator states $x$, contact events $k$, task labels $y$, and admission evidence $q$,
together with a ledger $\mathcal{L}$ that records every attempt and its outcome.

\subsection{Capability layer and task artifacts}
The farm separates what is shared across tasks from what is produced for each
task. The shared part is the capability layer $\mathcal{K}$, which comprises skills for
the stages of Section~\ref{sec:method}, role instructions, render templates that no
task may override, and the harnesses, ledgers, and verifiers that execute and check
the work; every run is pinned to a fixed version of $\mathcal{K}$. For each request,
the agent then generates the task artifacts
\begin{equation}
\mathcal{A}_\rho = \{\,G,\;T,\;S,\;C,\;J,\;R,\;B\,\}
\label{eq:artifacts}
\end{equation}
namely a grounding record $G$, a task contract $T$, a sampler $S$ of scenes and
initial states, an expert controller $C$, a judge specification $J$, a render contract
$R$, and the batch and dataset configurations $B$. This separation has two
consequences. First, a missing adapter is never a missing input, because the agent
produces every artifact itself; an artifact made for an earlier task serves only as an
example, and its thresholds, authorizations, and prompts are never loaded for a new
task. Second, behaviour that should generalize across tasks can change only in
$\mathcal{K}$, and only under tests. In addition, each fact in a delivered episode has exactly
one source that may certify it, so no agent role or program both proposes and certifies
the same fact (Appendix~\ref{sec:app-authority}).

\subsection{What does the farm cost?}
Replacing the teleoperator with an agent pays off only if the farm produces
useful data at a low total cost. Let $\eta$ be the fraction of attempts that
are admitted. For $N$ admitted episodes of one task, the farm then costs
\begin{equation}
C_{\mathrm{farm}}(N) = C_{\mathrm{adapt}} + C_{\mathrm{contract}}
 + \frac{N}{\eta}\left(C_{\mathrm{sim}}+C_{\mathrm{render}}+C_{\mathrm{review}}\right),
\end{equation}
where the one-time cost $C_{\mathrm{adapt}}$ covers grounding, simulator
qualification, and expert compilation, and $C_{\mathrm{contract}}$ is the render contract,
both including the agent's tokens and all failed attempts. Failed simulator and render attempts are not
free, since they enter through the yield $\eta$; what the farm removes is the physical
interaction, the hardware risk, and the manual reset of each attempt. Let
$m=(C_{\mathrm{sim}}+C_{\mathrm{render}}+C_{\mathrm{review}})/\eta$ be the farm's marginal cost
per admitted episode. The average cost per admitted episode,
\begin{equation}
\bar c(N)=\frac{C_{\mathrm{farm}}(N)}{N}=\frac{C_{\mathrm{adapt}}+C_{\mathrm{contract}}}{N}+m,
\label{eq:avgcost}
\end{equation}
falls towards $m$ as these two costs are spread over more episodes. Besides these
costs, the farm needs only the one calibration episode of the request; it is therefore
cheaper than teleoperation for $N$ episodes whenever a teleoperated episode costs more
than $(C_{\mathrm{adapt}}+C_{\mathrm{contract}}+Nm)/(N-1)$, provided that farm data trains policies at least as
good as the same number of teleoperated episodes.

\section{\ours}
\label{sec:method}

\ours turns a data request and one calibration episode into a stream of verified,
action-labelled episodes. Figure~\ref{fig:pipeline} follows the farm on the
pull-drawer task, and Algorithm~\ref{alg:farm} summarizes it for one task. We first
describe how the work is organized into stages that check one another
(Section~\ref{sec:overview}), and then the stages themselves: grounding the scene
(Section~\ref{sec:grounding}), qualifying the simulator (Section~\ref{sec:simqual}),
compiling the task and its expert (Section~\ref{sec:expert}), rendering
(Section~\ref{sec:render}), and admission (Section~\ref{sec:verify}).

\begin{figure*}[t]
  \centering
  \includegraphics[width=\textwidth]{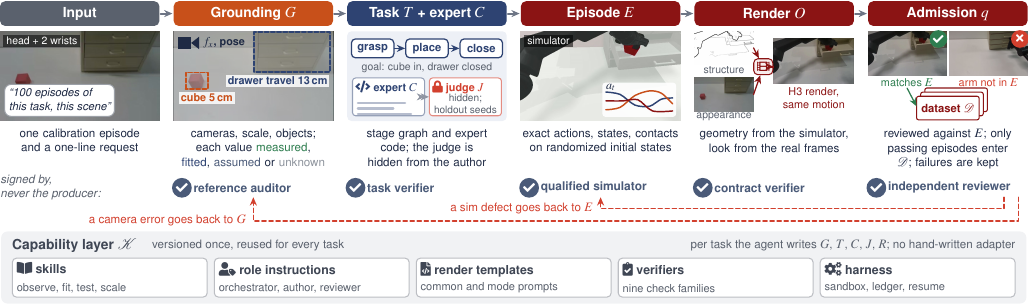}
  \caption{\textbf{The \ours pipeline, followed on one drawer episode.} The agent builds five
  intermediate representations: the grounding $G$, whose quantities are each labelled by how they are known;
  the task contract $T$ and the expert $C$, written against a hidden judge $J$; the simulated episodes $E$; the
  rendered observations $O$; and the admission record $q$. Each representation is checked by a component other
  than its producer (seals), and a rejection returns to the stage that caused it (dashed), as for the render here
  that shows an arm absent from $E$. The capability layer $\mathcal{K}$ is shared by all tasks; everything above
  it is written for this task.}
  \label{fig:method}
  \vspace{-1em}
\end{figure*}

\subsection{Overview: stages that check one another}
\label{sec:overview}
The farm compiles each task through a fixed sequence of intermediate
representations, much as a compiler lowers a program:
\begin{equation}
\adjustbox{max width=\linewidth}{$\displaystyle
e^c \xrightarrow{\text{ground}} G
\xrightarrow{\text{qualify, compile}} (T,C)
\xrightarrow[ z]{\text{rollout}} E
\xrightarrow[ p]{\text{render}} O
\xrightarrow{\text{verify}} (d,q).$}
\label{eq:compiler}
\end{equation}
The grounding $G$ describes the scene, the task contract $T$ and the expert $C$
define how the task is solved, the episode $E$ records one execution under a sampled
variation $z$, the observation $O$ is its rendering by a video model $p$, and the
admission record $q$ certifies the delivered episode $d$ (Figure~\ref{fig:method}).
Each representation is produced by one stage and accepted only after a component
other than its producer has checked it, because language models are unreliable judges
of their own output~\citep{huang2024selfcorrect}. A rejection is sent back to the earliest
stage responsible for the defect, and every failed attempt is kept, so the final
dataset comes with a complete record of how it was obtained. The stages are carried
out by coding agents~\citep{yang2024sweagent} in separate
roles~\citep{hong2024metagpt} that share one harness of skills, instructions,
templates, and verifiers. Because this harness is independent of any particular
task, the same farm adapts to a new task without changes to the harness; Appendix~\ref{sec:app-harness}
describes it in detail.

\subsection{Grounding from the calibration episode}
\label{sec:grounding}
\farmalgorithm
Grounding turns the calibration episode into a description of the scene that the
simulator can reproduce. From the full episode, the agent identifies the robot and
the objects, estimates their geometry and
articulation~\citep{chen2024urdformer,dai2024digitalcousins}, and fits the cameras and the
metric scale, cross-checking the scale against robot dimensions, known objects, or
multiple views. Physical quantities that a short video cannot determine, such as mass
or friction, are not guessed: every quantity is labelled as measured, fitted, assumed,
or unknown, and uncertain ones are covered by randomization in the
simulator~\citep{tobin2017domain,peng2018sim} rather than fixed to a confident but unsupported value. The agent also selects frames of the
calibration episode for every camera as appearance references for rendering, so the
user never has to supply images that the episode already contains.

\farmalgorithmpre
\subsection{Simulator qualification}
\label{sec:simqual}
\farmalgorithmwrap
Before the agent writes an expert, it checks that the simulator can reproduce the
skills the task requires, since no expert can succeed in a simulator that cannot, and
the failures of such a simulator would otherwise be attributed to the expert. The farm
therefore runs short probes of these skills over candidate simulator variants that adjust
material, contact, or actuator parameters, and selects the variant with the fewest
changes that passes every probe. The selection is issued as a certificate that binds
the variant's files and the result of every probe, the qualification script re-checks
the certificate against these results, and all episodes of the task are produced on
this certified simulator. Our implementation provides such probes for cloth; other
skills need probes of their own.

\subsection{Task contract and expert compilation}
\label{sec:expert}
The agent then specifies the task and writes an expert that solves it. Earlier
systems let a language model write the task together with its success check or
reward~\citep{wang2023gensim,wang2023robogen,ma2023eureka}; the farm instead prevents
the expert from fitting its own checks, a form of reward
hacking~\citep{skalse2022reward}. The task contract $T$
lists the stages of the task, the conditions that mark their completion, and the
shortcuts that are forbidden, such as teleporting objects or relaxing the success
criterion. A separate role turns the contract into a judge, which is frozen before
the expert is written.

The expert is then developed in a sandboxed harness that is shared by all tasks. The
author writes a controller as a program~\citep{liang2023codeaspolicies}, requests rollouts on development seeds, and receives the
judge's verdict on each, but it can neither read the judge nor run the held-out seeds
on which the expert is finally admitted; the harness alone executes rollouts and
records every attempt and its cost. The author may explore
freely during development, for example by trying alternative grasps on development
seeds, but such trials are search rather than demonstrations: an
admitted trajectory must be a single continuous execution, run afresh by the harness,
that passes the judge. Each produced episode samples the task's object and sensor
variation, by default including small changes in the height and tilt of the table, and
records the occupancy of every
joint's range so that the labels remain executable on the real robot.

\subsection{Rendering with real images}
\label{sec:render}
Rendering turns each simulated episode into photographic video while leaving its
actions, objects, and timing unchanged. The data either reproduce the recorded scene
(\textsc{sim2real}) or vary its appearance along axes the request authorizes
(\textsc{ood}), such as the tabletop material; an authorized axis never extends to
the robot, the number of objects, or the camera layout. For each task, the agent
writes a render contract that states the target appearance, the permitted
variations, and real reference frames for every camera. The contract is assembled
from templates that the agent cannot override, and the request actually sent to the
video model is checked against it before submission.
\vspace{-1em}

\paragraph{Rendering with MiniMax-H3.}
Our default renderer is MiniMax-H3 in its reference-to-video mode, with the accelerated
Sol-H3~\citep{li2026solh3} deployment. Four designs, none specific to a task, make its output
follow the simulator and keep the cameras consistent (Section~\ref{sec:res-views}); the
agent instantiates them for each task from the calibration episode.
\runin{Structure from the simulator, appearance from the real cameras.} As in
structure-conditioned video transfer~\citep{nvidia2025cosmostransfer,liu2025robotransfer},
edge maps of the simulator video drive a ControlNet~\citep{zhang2023controlnet} that fixes
the arm pose, the object positions, and the timing, while real frames and a picture of the
real robot supply the appearance.
\runin{All cameras in one generation.} The head and wrist views are tiled on one canvas and
generated as a single video, so they share lighting and materials and agree on which
objects exist and what each gripper holds.
\runin{References and prompts that add no content.} Reference frames match the simulated
task state, and each text prompt describes only its own segment, so the model neither
re-inserts content from its references nor draws later events too early.
\runin{Long episodes without drift.} Long episodes are generated as independent chunks and
joined where neighbouring chunks differ least, which avoids the drift of extending each
chunk from the previous output.
A vision-language model, run by the agent, compares the renders of each chunk with the
simulator video. Our recipe renders several seeds per chunk and keeps the one it selects;
the runs costed in Appendix~\ref{sec:app-cost} instead rendered each chunk once and
rendered again only the chunks it flagged.
\vspace{-1em}

\paragraph{Rendering with Seedance.}
The hosted Seedance API~\citep{gao2025seedance} is used through the same contract. We arrange the named
views in a fixed panel or render them separately, keep the full simulated motion
instead of retiming it, and split long episodes at simulator events. For either
renderer, bulk production begins with a small canary set that covers the hardest
occlusions and events.

\subsection{Verification, independent review, and admission}
\label{sec:verify}
An episode enters the dataset only if an independent reviewer accepts it after
automatic checks. The verifier is a single program shared by all tasks and
configured by a task-specific specification that the agent writes. It checks that the
rendered video contains the right objects and no extra ones, that events happen when
the simulation says they do, that the video is temporally stable and continuous across
chunks, that the background is static, that the appearance matches the real scene,
and that no layout has been copied from a reference frame. Before a specification is
used, it is calibrated, in the manner of mutation testing~\citep{jia2011mutation}, on
deliberately corrupted versions of an accepted episode, with
objects deleted, duplicated, or enlarged, events shifted in time, views swapped, or
colours changed, and every corruption must be caught by the check responsible for it.
Every task has its own verifier specification; the default H3 route also relies
on a frozen render evaluation and the seed selection of Section~\ref{sec:render}.

Automatic checks are necessary but not sufficient, because a generated video can look
plausible frame by frame and still contradict its labels~\citep{deng2026rbench}. Every episode is therefore
also examined by an independent reviewer~\citep{zheng2023judging,zhang2026veritas},
a separate agent that compares the decoded
video with the real reference frames and the simulated state and records its evidence
for each verdict. A final delivery check confirms that the dataset is complete,
unmodified, backed by the full record of attempts, and reviewed episode by episode;
the automatic checks gate the render stage rather than delivery. Policy training starts
only from a dataset whose admission record has been verified.

\section{Experimental Setup}
\label{sec:experiments}

\paragraph{Tasks.}
We evaluate \ours on five bimanual manipulation tasks on a dual-arm PiPER robot,
chosen so that each stresses a different part of the farm (Table~\ref{tab:tasks}):
nesting three rigid cups, placing an apple on a digital scale whose display changes
with the load, beating a drum in rhythm, putting a block into a drawer and closing it,
and folding a towel. For each task, the farm receives one calibration episode recorded
with a head camera and two wrist cameras together with a request for 100 episodes, and
the agent produces every task artifact from these inputs. Rigid and articulated
objects are simulated in SAPIEN through RoboTwin~2.0~\citep{chen2025robotwin2} and the
towel in Isaac Sim with particle cloth.
 Unless stated otherwise, observations are
rendered with our self-hosted MiniMax-H3 deployment, and Seedance serves as a second
renderer (Sections~\ref{sec:res-views} and~\ref{sec:res-hard}).

\begin{table*}[t]
\centering
\footnotesize
\setlength{\tabcolsep}{4pt}
\resizebox{1.0\textwidth}{!}{%
\begin{tabular}{lllll}
\toprule
Task & What makes it hard & Simulator & Success condition & Real-robot eval. \\
\midrule
Stack cups  & nesting thin-walled rigid cups             & SAPIEN    & all three cups nested in one stack       & ID, OOD \\
Weigh apple & placement on a scale with a live display   & SAPIEN    & apple at rest on the scale pan           & ID, OOD \\
Beat drum   & rhythmic contact timing                    & SAPIEN    & drum struck at least twice   & ID, OOD \\
Pull drawer & articulated object; pick, place, and close & SAPIEN    & block inside the drawer, drawer closed   & ID, OOD \\
Fold towel  & deformable cloth                           & Isaac Sim & towel folded twice          & ID, OOD \\
\bottomrule
\end{tabular}%
}
\caption{Tasks. Each starts from one calibration episode with a head and two
wrist cameras and a request for 100 episodes; the agent generates the scene,
camera fit, expert, judge, and render contract. ID: original scene; OOD: unseen
tablecloth and coloured light.}
\label{tab:tasks}
\end{table*}

\paragraph{Operator.}
The agent is a frontier coding model that operates the capability layer of
Section~\ref{sec:method}; in our runs, it can be GPT-6~Astra, GPT-5.6-Sol, or Claude Opus~5.5.

\paragraph{Policies and baselines.}
For every task, we post-train $\pi_{0.5}$~\citep{black2025pi05} on three
training sets of equal size $N=100$ that share the observation and action spaces, the
camera views, and the training recipe (50-step action horizon, batch size 128, 50k
steps), so that the sets differ only in where their episodes come from. \emph{Real
teleop} uses $N$ teleoperated episodes of the task in the original scene. \emph{Sim +
DR} is a conventional real-to-sim-to-real baseline that renders the same simulated
trajectories as ours directly in the simulator, with RoboTwin~2.0-style domain
randomization of textures, lighting, and distractors~\citep{chen2025robotwin2};
because it shares the grounding, the expert, and the actions with \ours, it isolates
the effect of how observations are produced. \emph{\ours}, finally, uses only the
farm's rendered episodes, without any real data. Its training set for each task mixes
renders of the original scene with renders whose tabletop material and lighting the
render contract varies through text, such as a wooden or stone tabletop under warmer
or cooler light; the tablecloths of the OOD evaluation appear in no training set.

\paragraph{Real-robot evaluation.}
Each policy is evaluated in 20 trials per task and condition, with initial
object placements randomized within the workspace and shared across policies, and a
trial counts as a success if it reaches the success condition of
Table~\ref{tab:tasks}. In-distribution (\emph{ID}) trials use the original scene,
whereas out-of-distribution (\emph{OOD}) trials cover the white table with a tablecloth that appears in no training
set (dark grey felt for stack cups and weigh apple, green gingham for the other three
tasks) and change the colour of the room light during the trial. We report the number
of successes for every task and condition.

\section{Results}
\label{sec:results}

We organize the results around five questions. Do policies trained only on farm
data work on the real robot, and how do they compare with teleoperated data and with a
conventional real-to-sim-to-real pipeline (Section~\ref{sec:res-policy})? What does the
farm cost (Section~\ref{sec:res-cost})? Do its episodes keep the
simulated motion that the action labels describe (Section~\ref{sec:res-motion}), and do
all three camera views both follow the simulation and look like real recordings, which a
direct call to the video model does not achieve (Section~\ref{sec:res-views})? And what does
video rendering add for objects that a simulator models poorly (Section~\ref{sec:res-hard})?

\subsection{Real-robot policies}
\label{sec:res-policy}

\begin{figure*}[t]
  \centering
  \includegraphics[width=\textwidth]{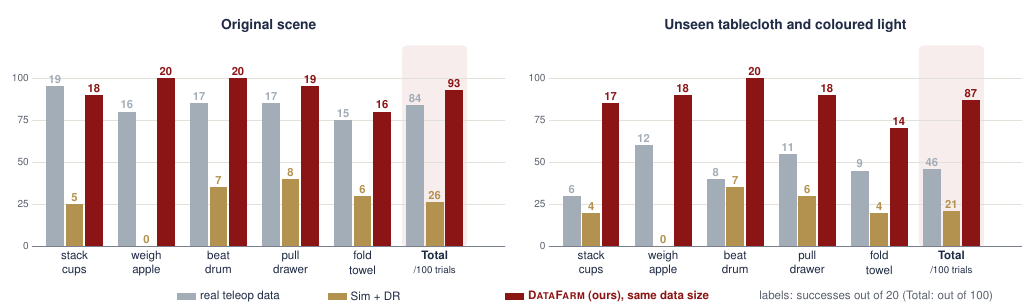}\\[3pt]
  \includegraphics[width=\textwidth]{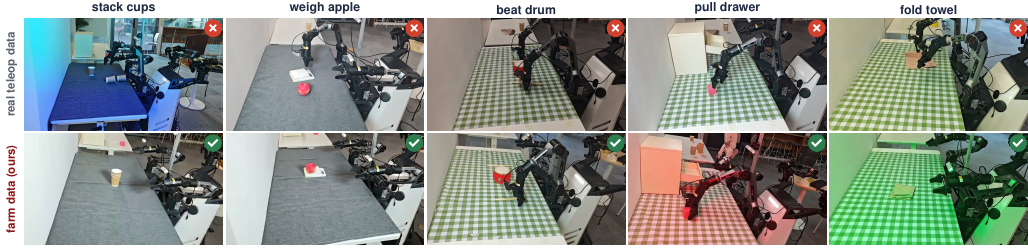}
  \caption{\textbf{Real-robot results.} \textbf{(a)}~Successes of $\pi_{0.5}$ policies trained
  on 100 episodes from each source, in 20 trials per task and condition (Total: out of 100), in the original
  scene (left) and under an unseen tablecloth and coloured light (right).
  \textbf{(b)}~Unseen-scene trials in which the policy trained on teleoperated data fails (top) and the policy
  trained only on farm data succeeds (bottom).}
  \label{fig:realrobot}
\end{figure*}

Policies trained only on farm data perform on par with teleoperated data in the
original scene and substantially better under unseen appearance
(Figure~\ref{fig:realrobot}a). In the original scene, they reach 93 successes in 100
trials against 84 for teleoperated data, matching or exceeding them on four of the
five tasks and falling one success behind on stack cups. Under OOD conditions, the
teleoperation-trained policies drop to 46 successes, whereas the farm-trained
policies keep 87 and lead on every task.
Figure~\ref{fig:realrobot}(b) shows typical OOD trials.

The conventional real-to-sim-to-real pipeline, by contrast, transfers poorly.
Policies trained on Sim + DR succeed in only 26 of 100 trials in the original scene
and 21 in the unseen scene, and they never complete weigh apple. Because Sim + DR
shares the grounding, the expert, and the actions with \ours and differs only in how
observations are produced, the gap of 93 against 26 and 87 against 21 successes
isolates the contribution of rendering with real images.

We attribute the robustness of farm-trained policies to the way the farm produces
data. Neither training set contains the test tablecloths or the coloured light, but
the renderer varies surfaces and lighting around fixed, exact trajectories at no
additional cost, so the policy cannot rely on how the table looks; a teleoperated
dataset, in contrast, shows only the scene in which it was recorded.

\subsection{Cost and fully automatic production}
\label{sec:res-cost}

\newcommand{\costfigcaption}{\textbf{What does a farm episode cost?} \textbf{(a)}~Average cost per admitted episode,
  $\bar c(N)=(C_{\mathrm{adapt}}+C_{\mathrm{contract}})/N+m$, for $N$ episodes of one task, mean over tasks, falling towards the marginal cost $m$ (dashed).
  Flat lines: teleoperation at the farm's mean episode length.
  \textbf{(b)}~One-time agent cost to five admitted simulated episodes, mean over three tasks (Table~\ref{tab:agents}, Appendix~\ref{sec:app-cost}).}
\ifwidelayout\else
\begin{figure}[t]
  \centering
  \includegraphics[width=\colfigwidth]{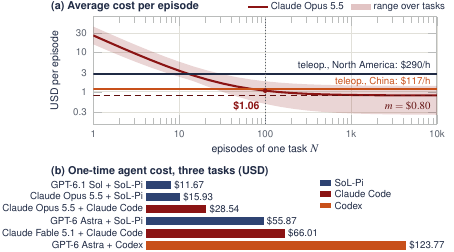}
  \caption{\costfigcaption}
  \label{fig:cost}
\end{figure}
\fi

\begin{table}[t]
\centering
\footnotesize
\setlength{\tabcolsep}{3pt}
\resizebox{0.8\linewidth}{!}{%
\begin{tabular}{lrrrrr}
\toprule
Stage & Stack cups & Weigh apple & Beat drum & Pull drawer & Fold towel \\
\midrule
One-time: grounding to admission (\S\ref{sec:grounding}--\ref{sec:expert}) & 18.22 & 10.33 & 19.25 & 19.40 & 37.38 \\
100 rollouts and audit                      & 7.43  & 8.34  & 7.66  & 9.31 & 13.41 \\
Render contract (\S\ref{sec:render})$^\S$               & 6.11  & 4.72  & 4.11  & 4.49  & 6.23 \\
\midrule
Total                         & 31.76 & 23.39 & 31.02 & 33.20 & 57.02 \\
\bottomrule
\end{tabular}%
}
\caption{Agent cost of each task, in USD at standard API pricing, by stage of
Section~\ref{sec:method}: from the calibration episode to 100 simulated episodes and a
render contract. One-time: grounding, simulator qualification, task contract, expert, and
admission of five simulated episodes, the scope of Figure~\ref{fig:cost}(b). Rollouts and
audit are charged per episode, and the render contract separately
(Table~\ref{tab:costmodel}). Task contracts and experts are written by Claude Opus~5.5, in SoL-Pi~\citep{liu2026solpi}.
$^\S$Excluding canary renders.}
\label{tab:cost}
\end{table}

\newcommand{\costmodeltabular}{%
\begin{tabular}{lrrrrr@{\hspace{12pt}}rr}
\toprule
 & \multirow{2}{*}{One-time} & \multirow{2}{*}{Contract} & \multirow{2}{*}{Per episode $m$} & \multicolumn{2}{c}{Average $\bar c(N)$} & \multicolumn{2}{c}{Per hour of data} \\
\cmidrule(lr){5-6}\cmidrule(lr){7-8}
Task & & & & $N=100$ & $N=1000$ & $m$ & 1\,h/task \\
\midrule
Stack cups & 18.22 & 6.11 & 0.93 & 1.17 & 0.96 & 77 & 101 \\
Weigh apple & 10.33 & 4.72 & 0.89 & 1.04 & 0.91 & 102 & 117 \\
Beat drum & 19.25 & 4.11 & 0.26 & 0.49 & 0.28 & 83 & 106 \\
Pull drawer & 19.40 & 4.49 & 0.52 & 0.76 & 0.54 & 55 & 79 \\
Fold towel & 37.38 & 6.23 & 1.41 & 1.84 & 1.45 & 88 & 132 \\
\midrule
Mean & 20.92 & 5.13 & 0.80 & 1.06 & 0.83 & 81 & 107 \\
\midrule
Teleop., N.\ America & -- & -- & 2.86 & 2.86 & 2.86 & 290 & 290 \\
Teleop., China, quote & -- & -- & 1.16 & 1.16 & 1.16 & 117 & 117 \\
\bottomrule
\end{tabular}}
\newcommand{\costmodelcaption}{Cost of the farm (USD). One-time and Contract: the one-time and render-contract rows of Table~\ref{tab:cost}.
Per episode $m$: rollouts and audit, H3 rendering, re-rendering of flagged chunks, and review per admitted episode, the same in the
original scene and under new appearance. Average: $\bar c(N)=(C_{\mathrm{adapt}}+C_{\mathrm{contract}})/N+m$ for $N$ episodes of one task.
Per hour of data, synchronized three-view episodes: $m$ times 3600 over the episode length, and the cost of
collecting one hour per task, one-time and contract costs included. Teleop.: labour in North America, estimated with Claude Opus 5.5, without
robot, venue, or setup; a full-service quote in China; per episode at the farm's mean length (Appendix~\ref{sec:app-cost}).}
\ifwidelayout
\begin{figure}[t]
\centering
\begin{minipage}[t]{0.5\textwidth}\vspace{0pt}\centering
  \includegraphics[width=\linewidth]{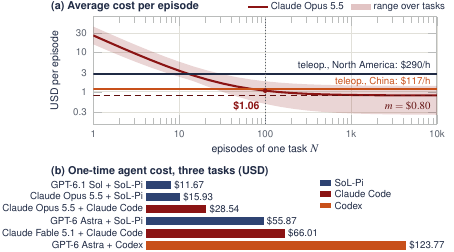}
  \captionof{figure}{\costfigcaption}\label{fig:cost}
\end{minipage}\hfill
\begin{minipage}[t]{0.48\textwidth}\vspace{0pt}\centering
  {\setlength{\tabcolsep}{3pt}\resizebox{\linewidth}{!}{\costmodeltabular}}
  \captionof{table}{\costmodelcaption}\label{tab:costmodel}
\end{minipage}
\end{figure}
\else
\begin{table}[t]
\centering
\footnotesize
\setlength{\tabcolsep}{3pt}
\resizebox{1.0\coltabwidth}{!}{\costmodeltabular}
\caption{\costmodelcaption}
\label{tab:costmodel}
\end{table}
\fi

The one-time cost of adapting a task depends on the agent. We tested six combinations of
four frontier language models and three coding-agent harnesses on three tasks, from the
calibration episode to five independently admitted simulated episodes
(Figure~\ref{fig:cost}(b), Table~\ref{tab:agents}). All six reach admission, at \$12--124
per task on average, and the differences mostly follow token prices. Cost is not the whole
comparison, however: the cheapest run needed follow-up instructions on two tasks and
admitted episodes that exceed the robot's joint speed limit (Appendix~\ref{sec:app-cost}).
We report the remaining costs with Claude Opus 5.5 in SoL-Pi~\citep{liu2026solpi}, which needed no human intervention. With these, the one-time cost is \$21 of API usage on average over the five
tasks, and a render contract adds \$5 (Table~\ref{tab:cost}). Every further episode needs
simulation, rendering, and review, but no robot, operator, or reset.

After adaptation, each admitted episode costs \$0.80 on average for rollouts and their
audit, rendering on rented GPUs, re-rendering of flagged chunks, and independent review. This cost is the same under
new appearance, which is a text change for the farm rather than a restaged scene. With the
one-time and contract costs spread over 100 episodes of a task, an episode costs \$1.06 on
average, and over 1{,}000 episodes \$0.83 (Figure~\ref{fig:cost}(a), Table~\ref{tab:costmodel}).
One hour of synchronized three-view data for a task costs \$107 on average, one-time and
contract costs included, and each further hour costs \$81.

These costs are lower than those of teleoperation. In North America, the labour of
teleoperated collection alone comes to about \$290 per usable hour of data by our
calculation with Claude Opus~5.5 (\$190--520 across the assumptions in
Appendix~\ref{sec:app-cost}), before any robot, venue, or setup cost, whereas the farm's
figures include its rented GPUs; the farm is cheaper after 10--37 episodes of a task. In
China, a quote we obtained for custom teleoperated data is about RMB 800 (\$117) per hour,
in line with reported prices~\citep{yicai2026mifeng}. This
full-service price is slightly above what the farm costs for the first hour of data on average, and the farm
becomes cheaper after 41--220 episodes of a task. The farm's costs rely on a self-hosted
renderer: at Seedance's published price, rendering one episode would cost \$12--63.

\newcommand{\flowcaption}{\textbf{Do renders keep the simulated motion?} Optical-flow error and share of
moving pixels whose flow agrees, between each render and its untextured simulator video (head view). Sim + DR
has exactly the simulated motion; the shifted H3 render is paired with the simulator video 0.5\,s later.
$^\dagger$H3 renders another towel episode, compared with its own simulator video.}
\ifwidelayout\Needspace{16\baselineskip}\fi
\subsection{Do renders keep the simulated motion?}
\label{sec:res-motion}
\ifwidelayout
\begin{wrapfigure}{r}{0.5\textwidth}\vspace{-2em}\centering
\includegraphics[width=\linewidth]{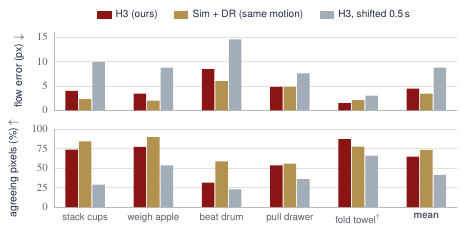}
\caption{\flowcaption}\label{fig:flow}\vspace{-1em}
\end{wrapfigure}
\else
\begin{figure}[t]
\centering
\includegraphics[width=\columnwidth]{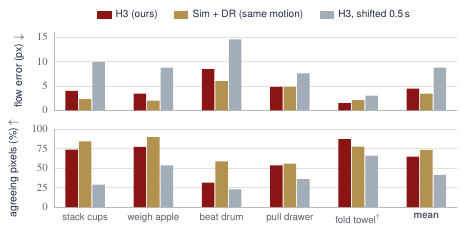}
\caption{\flowcaption}
\label{fig:flow}
\end{figure}
\fi
Action labels remain valid only if the rendered video shows the motion that the
simulator executed. We test this by estimating optical flow~\citep{teed2020raft} on each
render and on its untextured simulator video at the same frames (at 512$\times$288) and
comparing the two flow fields on moving pixels, by their mean difference and by the share
of pixels on which they agree (Figure~\ref{fig:flow}). Averaged over the five tasks, the
flow of H3 renders deviates from the simulated flow by 4.4 pixels and agrees on 65\% of
moving pixels. This is close to Sim + DR, which reproduces the simulated motion exactly
and still differs by 3.4 pixels, with 73\% agreement, only because the flow estimate
depends on appearance. Pairing the same H3 renders with the simulator video half a second
later, in contrast, raises the error to 8.8 pixels and lowers agreement to 41\%, so the
measure clearly separates matched from mismatched motion. Beat drum yields the largest
errors for every source, including Sim + DR, because its thin drumsticks move fast; even
there, H3 stays much closer to the simulated motion than the shifted control (8.5
against 14.6 pixels).

\subsection{Do all three views follow the simulation and look real?}
\label{sec:res-views}

\begin{figure*}[t]
  \centering
  \includegraphics[width=\textwidth]{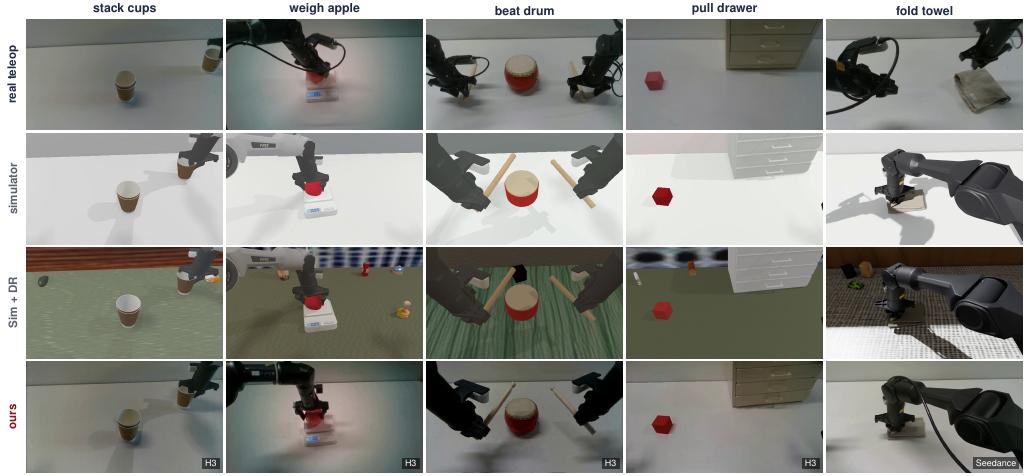}
  \caption{\textbf{Real, simulated, and farm episodes} at the same task phase (head view).
  Sim + DR adds domain-randomized textures and lighting to the simulator frame, and ours renders the simulator
  frame (renderer in the corner). Real and simulated frames come from different episodes, so object placements
  differ.}
  \label{fig:appearance}
\end{figure*}

\begin{figure*}[t]
  \centering
  \includegraphics[width=\textwidth]{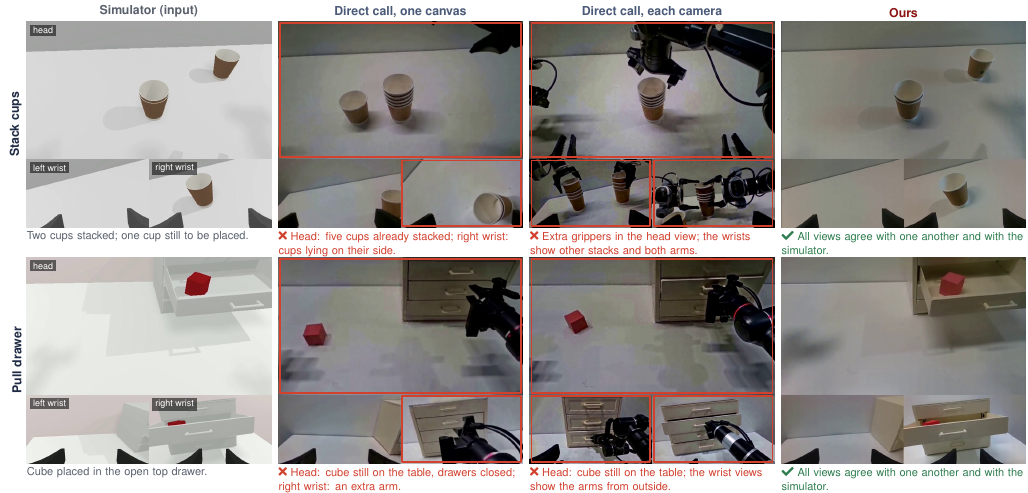}
  \caption{\textbf{Following the simulation in every view.} The same simulated frame rendered by calling the video model
  directly, with all cameras on one canvas or with each camera separately, and by the farm's renderer. Each
  cell shows the head view (top) and the left and right wrist views (bottom). Red boxes mark the views that
  contradict the simulator or the other views of the same frame; the notes describe the contradiction.}
  \label{fig:views}
\end{figure*}

A policy reads the head and both wrist cameras together, so every view of a farm episode
must be correct in two respects at once: it must follow the simulated episode, whose
actions are the labels, and look like the real camera that the policy sees at deployment.
The simulator video meets only the first requirement, and a direct call to the video
model mostly only the second.

Farm episodes look far closer to real recordings than simulator renderings, both visually
and in image statistics. Figure~\ref{fig:appearance} shows the same task phase in a real
recording, in the simulator, with domain randomization, and in a farm episode. The
simulator and its randomized variant reproduce the geometry but not the materials, the
lighting, or the robot's appearance, whereas the farm episode keeps the simulated geometry
and takes its appearance from the calibration episode. Image statistics confirm this
impression: averaged over the tasks that every source renders, the kernel distance (KID,
$\times 10^3$) of head-view frames to real camera frames is 0.83 for H3 and 0.88 for
Seedance, against 1.93 for the simulator and 2.50 with domain randomization, and the
farm is closer than the simulator on every task (Appendix~\ref{sec:app-kid}).

Following the simulation in every view is where calling the same video model directly
fails. We call MiniMax-H3 with the simulator video as motion reference, real reference frames,
and the contract prompt, on one canvas or per camera, and correlate the frame-difference
energy and image gradients of each view with the simulator video, on the episodes of stack
cups, weigh apple, and pull drawer on which our recipe was developed.

\newcommand{\viewstable}{\footnotesize\setlength{\tabcolsep}{3pt}%
\begin{tabular}{lcccc}
\toprule
Renderer & Motion $\uparrow$ & Structure $\uparrow$ & $|$Lag$|$ & Jumps \\
\midrule
Direct call, one canvas   & 0.07 & 0.16 & 12.7 & 11.2 \\
Direct call, each camera  & 0.04 & 0.10 & 14.7 & 8.8 \\
Edge control only         & 0.50 & 0.54 & 0    & 30.2 \\
Ours, each camera$^\dagger$ & 0.55 & 0.64 & 0    & 40.6 \\
Ours                      & \textbf{0.74} & \textbf{0.85} & \textbf{0} & 2.9 \\
\midrule
Seedance (hosted)         & 0.73 & 0.72 & 0.3  & 2.3 \\
\bottomrule
\end{tabular}}
\newcommand{\viewscaption}{Renderers on the same simulated episode of stack cups, weigh apple, and pull drawer.
Motion and structure: correlation with the simulator video of frame-difference energy and of image
gradients; lag in frames; jumps per 1{,}000 frames. Means over the three episodes.
$^\dagger$Stack cups only; ours on that episode: 0.75, 0.87, 0, 4.2.}
\ifwidelayout
\begin{wraptable}{r}{0.5\textwidth}\vspace{-1.2em}\centering
\resizebox{\linewidth}{!}{\viewstable}
\caption{\viewscaption}\label{tab:views}\vspace{-1em}
\end{wraptable}
\else
\begin{table}[!b]
\centering
\resizebox{1.0\coltabwidth}{!}{\viewstable}
\caption{\viewscaption}
\label{tab:views}
\end{table}
\fi
Calling the video model directly fails in two ways (Figure~\ref{fig:views} and
Table~\ref{tab:views}). On one canvas, it reproduces the task state of its reference
frames instead of the simulated motion: its motion correlation is 0.07, against 0.74 for
the farm's renderer, and its head view often shows a different task state than a wrist
view. Rendered separately, each camera is generated without knowledge of the others, so
the wrist views show extra arms, other cup stacks, or the robot from outside. Generating
all cameras together is therefore necessary: applied to each camera of stack cups
separately, even the farm's renderer invents cups in the wrist views, and its motion and
structure correlations fall from 0.75 and 0.87 to 0.55 and 0.64.
Edge control alone makes the render follow the simulator (motion 0.50) but leaves 30
jumps per 1{,}000 frames, which the references, prompts, and chunk assembly of
Section~\ref{sec:render} reduce to 2.9.

Of the self-hosted renderers, only ours meets both requirements in all three views, on
par with the hosted Seedance (KID 0.83 against 0.88; motion 0.74 against 0.73), so the
labels describe what every camera sees.

\newcommand{\hardcaption}{\textbf{Appearance that the simulator does not model.} Crops of a simulator frame
  (top) and of the farm episode rendered from that frame (middle), with the same object in a real recording at
  the same task phase (bottom; not the same frame). The simulator draws the scale's display as a flat
  texture, the drum as a red cylinder, and the towel as smooth particle cloth; the farm episode shows a lit
  display with the simulated reading, a lacquered drum with brass tacks, and the weave of the real towel.}
\ifwidelayout\Needspace{6\baselineskip}\fi
\subsection{Objects that are hard to simulate}
\label{sec:res-hard}
\ifwidelayout
\begin{figure}[!t]
  \centering
  \begin{minipage}[c]{0.64\textwidth}
  \includegraphics[width=\linewidth]{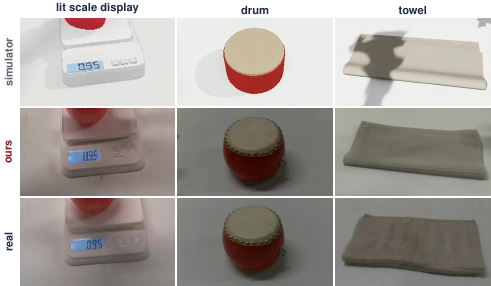}
  \end{minipage}\hfill
  \begin{minipage}[c]{0.33\textwidth}
  \caption{\hardcaption}\label{fig:hard}
  \end{minipage}
\end{figure}
\else
\begin{figure}[t]
  \centering
  \includegraphics[width=\columnwidth]{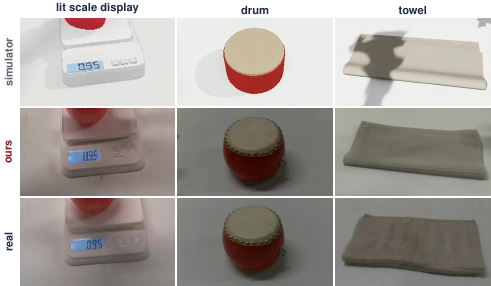}
  \caption{\hardcaption}
  \label{fig:hard}
\end{figure}
\fi

Some objects are hard to simulate because of how they look rather than how they move.
For a scale's display, a drum, and a towel, the farm episode is much closer to the real
recording than to the simulator (Figure~\ref{fig:hard}).

This realism is also cheap. A conventional real-to-sim-to-real pipeline would have to
model each of these appearances, with an emissive display, a scanned drum, and a
texture that deforms with the cloth. The farm needs only the geometry and contact that
the physics requires and takes appearance from the calibration episode, set once per
task by a render contract costing \$4.1--6.2 (Table~\ref{tab:cost}). Motion and object
states still come from the simulator (Section~\ref{sec:res-motion}).

\section{Limitations, Ethics, and Broader Impact}
\label{sec:limitations}

\ours depends on a simulator that can reproduce the skills of each task.
Simulator qualification (Section~\ref{sec:simqual}) detects when the available
simulator cannot, but such a task then needs a new simulator variant before the farm
can produce data for it. Similarly, a calibration episode cannot identify dynamics
that it does not exhibit, such as mass or friction; the farm labels these quantities
as assumed and randomizes over them rather than recovering them, which may not
suffice for tasks whose success depends on precise dynamics. The verification of
generated video is also imperfect: rendered episodes can contain defects that
automatic checks miss, which is why every episode undergoes independent review and
why rejected renders count toward the cost. Finally, our
evaluation covers one dual-arm embodiment and five tasks, and other embodiments and
task families remain to be tested.

Because farm episodes are photographic videos of real scenes, their provenance must
remain traceable. Each episode therefore carries a record that binds it to its
simulated state, renderer, references, and review, so that a dataset can always be
identified as generated. Cheaper data, moreover, does not make deployment safer: a
policy trained on farm data still requires real-robot evaluation before use.

\section{Conclusion}
\label{sec:conclusion}

We have shown that a coding agent can take the teleoperator's place in
collecting robot data. Given a request and one calibration episode of the scene, the
agent compiles the task into a qualified simulation, an expert, and a render
contract, and the farm turns this compilation into verified, real-ready episodes at
scale, on which policies are trained directly and deployed on the real robot. Correctness does not rest on trusting the agent, because hidden judges, the
qualified simulator, and independent reviewers certify what it produces, and every
failed attempt is recorded. On the real robot, policies trained only on farm data
match policies trained on the same number of teleoperated episodes, hold up far
better under appearance that neither training set contained, and far outperform a
conventional real-to-sim-to-real pipeline. The farm pays a one-time agent cost of \$21 per task on
average, \$5 for a render contract, and \$0.80 for every further episode, without a robot, an operator, or a reset;
one hour of three-view data for a task costs \$107, these costs included,
slightly below a quote of \$117 for teleoperation in China and below \$290
in labour alone in North America.

\clearpage
\bibliographystyle{plainnat}
\setlength{\bibhang}{0pt}
\setlength\bibindent{0pt}
\bibliography{references}

\begin{thebibliography}{61}
\providecommand{\natexlab}[1]{#1}
\providecommand{\url}[1]{\texttt{#1}}
\expandafter\ifx\csname urlstyle\endcsname\relax
  \providecommand{\doi}[1]{doi: #1}\else
  \providecommand{\doi}{doi: \begingroup \urlstyle{rm}\Url}\fi

\bibitem[Ahn et~al.(2024)Ahn, Dwibedi, Finn, et~al.]{ahn2024autort}
Michael Ahn, Debidatta Dwibedi, Chelsea Finn, et~al.
\newblock Autort: Embodied foundation models for large scale orchestration of
  robotic agents.
\newblock \emph{arXiv preprint arXiv:2401.12963}, 2024.

\bibitem[Black et~al.(2024)Black, Brown, Driess, et~al.]{black2024pi0}
Kevin Black, Noah Brown, Danny Driess, et~al.
\newblock $\pi_0$: A vision-language-action flow model for general robot
  control.
\newblock \emph{arXiv preprint arXiv:2410.24164}, 2024.

\bibitem[Black et~al.(2025)Black, Brown, Darpinian, et~al.]{black2025pi05}
Kevin Black, Noah Brown, James Darpinian, et~al.
\newblock $\pi_{0.5}$: A vision-language-action model with open-world
  generalization.
\newblock \emph{arXiv preprint arXiv:2504.16054}, 2025.

\bibitem[Chen et~al.(2026{\natexlab{a}})Chen, Liu, Sukhatme, and
  Seita]{chen2026craft}
Jason Chen, I-Chun~Arthur Liu, Gaurav Sukhatme, and Daniel Seita.
\newblock Craft: Video diffusion for bimanual robot data generation.
\newblock \emph{arXiv preprint arXiv:2604.03552}, 2026{\natexlab{a}}.

\bibitem[Chen et~al.(2026{\natexlab{b}})Chen, Lai, Wang, Ouyang, Huang, Yue,
  et~al.]{chen2026mcsparse}
Jiarui Chen, Zeqiang Lai, Jiangshan Wang, Ziheng Ouyang, Ye~Huang, Xiangyu Yue,
  et~al.
\newblock {MC-Sparse}: Deconstructing and closing the dense-sparse attention
  gap in diffusion transformers.
\newblock \emph{arXiv preprint arXiv:2610.06801}, 2026{\natexlab{b}}.

\bibitem[Chen et~al.(2025)Chen, Chen, Chen, et~al.]{chen2025robotwin2}
Tianxing Chen, Zanxin Chen, Baijun Chen, et~al.
\newblock Robotwin 2.0: A scalable data generator and benchmark with strong
  domain randomization for robust bimanual robotic manipulation.
\newblock \emph{arXiv preprint arXiv:2506.18088}, 2025.

\bibitem[Chen et~al.(2023)Chen, Kiami, Gupta, and Kumar]{chen2023genaug}
Zoey Chen, Sho Kiami, Abhishek Gupta, and Vikash Kumar.
\newblock Genaug: Retargeting behaviors to unseen situations via generative
  augmentation.
\newblock In \emph{Robotics: Science and Systems (RSS)}, 2023.

\bibitem[Chen et~al.(2024)Chen, Walsman, Memmel, et~al.]{chen2024urdformer}
Zoey Chen, Aaron Walsman, Marius Memmel, et~al.
\newblock Urdformer: A pipeline for constructing articulated simulation
  environments from real-world images.
\newblock \emph{arXiv preprint arXiv:2405.11656}, 2024.

\bibitem[Chi et~al.(2024)Chi, Xu, Pan, Cousineau, Burchfiel, Feng, Tedrake, and
  Song]{chi2024umi}
Cheng Chi, Zhenjia Xu, Chuer Pan, Eric Cousineau, Benjamin Burchfiel, Siyuan
  Feng, Russ Tedrake, and Shuran Song.
\newblock Universal manipulation interface: In-the-wild robot teaching without
  in-the-wild robots.
\newblock In \emph{Robotics: Science and Systems (RSS)}, 2024.

\bibitem[Dai et~al.(2024)Dai, Wong, Jiang, et~al.]{dai2024digitalcousins}
Tianyuan Dai, Josiah Wong, Yunfan Jiang, et~al.
\newblock Automated creation of digital cousins for robust policy learning.
\newblock \emph{arXiv preprint arXiv:2410.07408}, 2024.

\bibitem[Deng et~al.(2026)Deng, Pan, Zhang, et~al.]{deng2026rbench}
Yufan Deng, Zilin Pan, Hongyu Zhang, et~al.
\newblock Rethinking video generation model for the embodied world.
\newblock \emph{arXiv preprint arXiv:2601.15282}, 2026.

\bibitem[Fu et~al.(2026)Fu, Zhang, Gong, Deng, Sun, Min,
  et~al.]{fu2026stablevla}
Yiyang Fu, Chubin Zhang, Shukai Gong, Yufan Deng, Kaiwei Sun, Qiyang Min,
  et~al.
\newblock {StableVLA}: Towards robust vision-language-action models without
  extra data.
\newblock In \emph{International Conference on Machine Learning (ICML)}, 2026.

\bibitem[Gao et~al.(2025)Gao, Guo, Hoang, Huang, Jiang, Kong,
  et~al.]{gao2025seedance}
Yu~Gao, Haoyuan Guo, Tuyen Hoang, Weilin Huang, Lu~Jiang, Fangyuan Kong, et~al.
\newblock Seedance 1.0: Exploring the boundaries of video generation models.
\newblock \emph{arXiv preprint arXiv:2506.09113}, 2025.

\bibitem[Garrett et~al.(2024)Garrett, Mandlekar, Wen, and
  Fox]{garrett2024skillgen}
Caelan Garrett, Ajay Mandlekar, Bowen Wen, and Dieter Fox.
\newblock Skillmimicgen: Automated demonstration generation for efficient skill
  learning and deployment.
\newblock \emph{arXiv preprint arXiv:2410.18907}, 2024.

\bibitem[Gong et~al.(2026)Gong, Zhai, Zhang, Cui, Huang, Fu,
  et~al.]{gong2026vigar}
Shukai Gong, Xuanran Zhai, Yintianrun Zhang, Ruopeng Cui, Ye~Huang, Yiyang Fu,
  et~al.
\newblock Rethinking world-action model for compositional and in-context
  robotic manipulation.
\newblock \emph{arXiv preprint arXiv:2610.02368}, 2026.

\bibitem[Han et~al.(2025)Han, Liu, Chen, et~al.]{han2025re3sim}
Xiaoshen Han, Minghuan Liu, Yilun Chen, et~al.
\newblock Re$^3$sim: Generating high-fidelity simulation data via
  3d-photorealistic real-to-sim for robotic manipulation.
\newblock \emph{arXiv preprint arXiv:2502.08645}, 2025.

\bibitem[Han et~al.(2026)Han, Qiu, Bai, et~al.]{han2026video2sim2real}
Yunhai Han, Jianuo Qiu, Linhao Bai, et~al.
\newblock Video2sim2real: Full-stack autonomous dexterous skill acquisition
  from a single human video.
\newblock \emph{arXiv preprint arXiv:2606.08828}, 2026.

\bibitem[Hong et~al.(2024)Hong, Zhuge, Chen, Zheng, Cheng, Zhang, Wang, Wang,
  Yau, Lin, et~al.]{hong2024metagpt}
Sirui Hong, Mingchen Zhuge, Jonathan Chen, Xiawu Zheng, Yuheng Cheng, Ceyao
  Zhang, Jinlin Wang, Zili Wang, Steven Ka~Shing Yau, Zijuan Lin, et~al.
\newblock {MetaGPT}: Meta programming for a multi-agent collaborative
  framework.
\newblock In \emph{International Conference on Learning Representations
  (ICLR)}, 2024.

\bibitem[Hu et~al.(2024)Hu, Lu, and Clune]{hu2024adas}
Shengran Hu, Cong Lu, and Jeff Clune.
\newblock Automated design of agentic systems.
\newblock \emph{arXiv preprint arXiv:2408.08435}, 2024.

\bibitem[Huang et~al.(2024)Huang, Chen, Mishra, Zheng, Yu, Song, and
  Zhou]{huang2024selfcorrect}
Jie Huang, Xinyun Chen, Swaroop Mishra, Huaixiu~Steven Zheng, Adams~Wei Yu,
  Xinying Song, and Denny Zhou.
\newblock Large language models cannot self-correct reasoning yet.
\newblock In \emph{International Conference on Learning Representations
  (ICLR)}, 2024.

\bibitem[Huang et~al.(2026)Huang, Huang, Liao, Niu, Li, Zhou,
  et~al.]{huang2026dcsae}
Xu~Huang, Ye~Huang, Zijun Liao, Yuwei Niu, Xiaojie Li, Menghan Zhou, et~al.
\newblock {DC-SAE}: Deep compression semantic autoencoder for faster diffusion
  convergence.
\newblock In \emph{Advances in Neural Information Processing Systems
  (NeurIPS)}, 2026.

\bibitem[Jang et~al.(2025)Jang, Ye, Lin, et~al.]{jang2025dreamgen}
Joel Jang, Seonghyeon Ye, Zongyu Lin, et~al.
\newblock Dreamgen: Unlocking generalization in robot learning through video
  world models.
\newblock \emph{arXiv preprint arXiv:2505.12705}, 2025.

\bibitem[Jia and Harman(2011)]{jia2011mutation}
Yue Jia and Mark Harman.
\newblock An analysis and survey of the development of mutation testing.
\newblock \emph{IEEE Transactions on Software Engineering}, 37\penalty0
  (5):\penalty0 649--678, 2011.

\bibitem[Jiang et~al.(2025)Jiang, Xie, Lin, Xu, Wan, Mandlekar, Fan, and
  Zhu]{jiang2025dexmimicgen}
Zhenyu Jiang, Yuqi Xie, Kevin Lin, Zhenjia Xu, Weikang Wan, Ajay Mandlekar,
  Linxi Fan, and Yuke Zhu.
\newblock Dexmimicgen: Automated data generation for bimanual dexterous
  manipulation via imitation learning.
\newblock In \emph{IEEE International Conference on Robotics and Automation
  (ICRA)}, 2025.

\bibitem[Kim et~al.(2024)Kim, Pertsch, Karamcheti, et~al.]{kim2024openvla}
Moo~Jin Kim, Karl Pertsch, Siddharth Karamcheti, et~al.
\newblock Openvla: An open-source vision-language-action model.
\newblock \emph{arXiv preprint arXiv:2406.09246}, 2024.

\bibitem[Li et~al.(2024)Li, Li, Wang, et~al.]{li2024robogsim}
Xinhai Li, Jialin Li, Ruiping Wang, et~al.
\newblock Robogsim: A real2sim2real robotic gaussian splatting simulator.
\newblock \emph{arXiv preprint arXiv:2411.11839}, 2024.

\bibitem[Li et~al.(2026)Li, Yu, Chen, Li, Xue, Liu, et~al.]{li2026solh3}
Yitong Li, Jincheng Yu, Junsong Chen, Haopeng Li, Shuchen Xue, Haozhe Liu,
  et~al.
\newblock {Sol-H3}: Recursive self-improvement for {MiniMax-H3} inference
  acceleration on {Sol-Engine} across cloud and edge.
\newblock \emph{arXiv preprint arXiv:2609.35110}, 2026.

\bibitem[Liang et~al.(2022)Liang, Huang, Xia, et~al.]{liang2023codeaspolicies}
Jacky Liang, Wenlong Huang, Fei Xia, et~al.
\newblock Code as policies: Language model programs for embodied control.
\newblock \emph{arXiv preprint arXiv:2209.07753}, 2022.

\bibitem[Liu et~al.(2026{\natexlab{a}})Liu, Huang, Huang, Wen, Liu, Zhang,
  et~al.]{liu2026fvattn}
Hao Liu, Chenghuan Huang, Ye~Huang, Zhiying Wen, Hao Liu, Mohan Zhang, et~al.
\newblock {FVAttn}: Adaptive sparse attention with runtime load balancing for
  video generation.
\newblock \emph{arXiv preprint arXiv:2607.16190}, 2026{\natexlab{a}}.

\bibitem[Liu et~al.(2026{\natexlab{b}})Liu, Ye, Gao, Cao, Li, Zhuge,
  et~al.]{liu2026solpi}
Haozhe Liu, Tian Ye, Sensen Gao, Qihang Cao, Yitong Li, Mingchen Zhuge, et~al.
\newblock {SoL-Pi}: Recursively scaling auto-research loops for efficient agent
  harness.
\newblock \emph{arXiv preprint arXiv:2609.20519}, 2026{\natexlab{b}}.

\bibitem[Liu et~al.(2025)Liu, Wang, Zhao, Li, Qin, Qiu, Zhu, Huang, and
  Su]{liu2025robotransfer}
Liu Liu, Xiaofeng Wang, Guosheng Zhao, Keyu Li, Wenkang Qin, Jiaxiong Qiu,
  Zheng Zhu, Guan Huang, and Zhizhong Su.
\newblock Robotransfer: Geometry-consistent video diffusion for robotic visual
  policy transfer.
\newblock \emph{arXiv preprint arXiv:2505.23171}, 2025.

\bibitem[Ma et~al.(2026)Ma, Bi, Deng, Zhai, Zhang, Huang,
  et~al.]{ma2026humanscale}
Juncheng Ma, Jianxin Bi, Yufan Deng, Xuanran Zhai, Kewei Zhang, Ye~Huang,
  et~al.
\newblock {HumanScale}: Egocentric human video can outperform real-robot data
  for embodied pretraining.
\newblock \emph{arXiv preprint arXiv:2606.20521}, 2026.

\bibitem[Ma et~al.(2023)Ma, Liang, Wang, et~al.]{ma2023eureka}
Yecheng~Jason Ma, William Liang, Guanzhi Wang, et~al.
\newblock Eureka: Human-level reward design via coding large language models.
\newblock \emph{arXiv preprint arXiv:2310.12931}, 2023.

\bibitem[Mandlekar et~al.(2023)Mandlekar, Nasiriany, Wen,
  et~al.]{mandlekar2023mimicgen}
Ajay Mandlekar, Soroush Nasiriany, Bowen Wen, et~al.
\newblock Mimicgen: A data generation system for scalable robot learning using
  human demonstrations.
\newblock \emph{arXiv preprint arXiv:2310.17596}, 2023.

\bibitem[{NVIDIA}(2025)]{nvidia2025cosmostransfer}
{NVIDIA}.
\newblock Cosmos-transfer2.5: Conditional world generation for physical ai.
\newblock \url{https://research.nvidia.com/labs/dir/cosmos-transfer2.5/}, 2025.
\newblock Accessed 2026-09-19.

\bibitem[Peng et~al.(2018)Peng, Andrychowicz, Zaremba, and Abbeel]{peng2018sim}
Xue~Bin Peng, Marcin Andrychowicz, Wojciech Zaremba, and Pieter Abbeel.
\newblock Sim-to-real transfer of robotic control with dynamics randomization.
\newblock In \emph{IEEE International Conference on Robotics and Automation
  (ICRA)}, 2018.

\bibitem[Qureshi et~al.(2025)Qureshi, Garg, Yandun, Held, Kantor, and
  Silwal]{qureshi2025splatsim}
Mohammad~Nomaan Qureshi, Sparsh Garg, Francisco Yandun, David Held, George
  Kantor, and Abhisesh Silwal.
\newblock Splatsim: Zero-shot sim2real transfer of rgb manipulation policies
  using gaussian splatting.
\newblock In \emph{IEEE International Conference on Robotics and Automation
  (ICRA)}, 2025.

\bibitem[Ranawaka et~al.(2026)Ranawaka, Wong, Pai,
  et~al.]{ranawaka2026simfoundry}
Nadun Ranawaka, Josiah Wong, Wei-Lin Pai, et~al.
\newblock Simfoundry: Modular and automated scene generation for policy
  learning and evaluation.
\newblock \emph{arXiv preprint arXiv:2606.28276}, 2026.

\bibitem[{Robotics Center of Silicon Valley}(2026)]{rcsv2026cost}
{Robotics Center of Silicon Valley}.
\newblock How much does robot data collection cost in 2026?
\newblock \url{https://www.roboticscenter.ai/blog/robot-data-collection-cost},
  2026.

\bibitem[Si et~al.(2026)Si, Bi, Yang, Ni, Huang, Wang,
  et~al.]{si2026fewertokens}
Ruiyang Si, Jianxin Bi, Shunyu Yang, Rui Ni, Wenbo Huang, Qiang Wang, et~al.
\newblock Fewer tokens, better action: {GPT-6 Astra} robot agents with 14\%
  higher success rate but 65\% fewer tokens.
\newblock \emph{arXiv preprint arXiv:2610.01939}, 2026.

\bibitem[Skalse et~al.(2022)Skalse, Howe, Krasheninnikov, and
  Krueger]{skalse2022reward}
Joar Skalse, Nikolaus Howe, Dmitrii Krasheninnikov, and David Krueger.
\newblock Defining and characterizing reward hacking.
\newblock In \emph{Advances in Neural Information Processing Systems
  (NeurIPS)}, 2022.

\bibitem[Teed and Deng(2020)]{teed2020raft}
Zachary Teed and Jia Deng.
\newblock {RAFT}: Recurrent all-pairs field transforms for optical flow.
\newblock In \emph{European Conference on Computer Vision (ECCV)}, 2020.

\bibitem[Tobin et~al.(2017)Tobin, Fong, Ray, Schneider, Zaremba, and
  Abbeel]{tobin2017domain}
Josh Tobin, Rachel Fong, Alex Ray, Jonas Schneider, Wojciech Zaremba, and
  Pieter Abbeel.
\newblock Domain randomization for transferring deep neural networks from
  simulation to the real world.
\newblock In \emph{IEEE/RSJ International Conference on Intelligent Robots and
  Systems (IROS)}, 2017.

\bibitem[Torne et~al.(2024)Torne, Simeonov, Li, et~al.]{torne2024rialto}
Marcel Torne, Anthony Simeonov, Zechu Li, et~al.
\newblock Reconciling reality through simulation: A real-to-sim-to-real
  approach for robust manipulation.
\newblock \emph{arXiv preprint arXiv:2403.03949}, 2024.

\bibitem[Wang et~al.(2025)Wang, Meng, Wang, et~al.]{wang2025embodiedreamer}
Boyuan Wang, Xinpan Meng, Xiaofeng Wang, et~al.
\newblock Embodiedreamer: Advancing real2sim2real transfer for policy training
  via embodied world modeling.
\newblock \emph{arXiv preprint arXiv:2507.05198}, 2025.

\bibitem[Wang et~al.(2023{\natexlab{a}})Wang, Ling, Yuan,
  et~al.]{wang2023gensim}
Lirui Wang, Yiyang Ling, Zhecheng Yuan, et~al.
\newblock Gensim: Generating robotic simulation tasks via large language
  models.
\newblock \emph{arXiv preprint arXiv:2310.01361}, 2023{\natexlab{a}}.

\bibitem[Wang et~al.(2023{\natexlab{b}})Wang, Xian, Chen,
  et~al.]{wang2023robogen}
Yufei Wang, Zhou Xian, Feng Chen, et~al.
\newblock Robogen: Towards unleashing infinite data for automated robot
  learning via generative simulation.
\newblock \emph{arXiv preprint arXiv:2311.01455}, 2023{\natexlab{b}}.

\bibitem[Wei et~al.(2025)Wei, Huang, Zhou, Jiang, Du, and Lu]{wei2025ghidorah}
Jinhui Wei, Ye~Huang, Yuhui Zhou, Jiazhi Jiang, Jiangsu Du, and Yutong Lu.
\newblock Ghidorah: Fast {LLM} inference on edge with speculative decoding and
  hetero-core parallelism.
\newblock \emph{arXiv preprint arXiv:2505.23219}, 2025.

\bibitem[Xue et~al.(2025)Xue, Deng, Chen, et~al.]{xue2025demogen}
Zhengrong Xue, Shuying Deng, Zhenyang Chen, et~al.
\newblock Demogen: Synthetic demonstration generation for data-efficient
  visuomotor policy learning.
\newblock \emph{arXiv preprint arXiv:2502.16932}, 2025.

\bibitem[Yang et~al.(2024)Yang, Jimenez, Wettig, et~al.]{yang2024sweagent}
John Yang, Carlos~E. Jimenez, Alexander Wettig, et~al.
\newblock Swe-agent: Agent-computer interfaces enable automated software
  engineering.
\newblock \emph{arXiv preprint arXiv:2405.15793}, 2024.

\bibitem[{Yicai}(2026)]{yicai2026mifeng}
{Yicai}.
\newblock Report on the mifeng embodied-{AI} data platform.
\newblock
  \url{https://finance.sina.com.cn/tech/roll/2026-04-17/doc-inhuuimk6582970.shtml},
  2026.
\newblock 17 April 2026.

\bibitem[Yu et~al.(2025)Yu, Fu, Huang, El-Refai, Ambrus, Cheng, Irshad, and
  Goldberg]{yu2025real2render2real}
Justin Yu, Letian Fu, Huang Huang, Karim El-Refai, Rares~Andrei Ambrus, Richard
  Cheng, Muhammad~Zubair Irshad, and Ken Goldberg.
\newblock Real2render2real: Scaling robot data without dynamics simulation or
  robot hardware.
\newblock \emph{arXiv preprint arXiv:2505.09601}, 2025.

\bibitem[Yu et~al.(2023)Yu, Xiao, Stone, Tompson, Brohan, et~al.]{yu2023rosie}
Tianhe Yu, Ted Xiao, Austin Stone, Jonathan Tompson, Anthony Brohan, et~al.
\newblock Scaling robot learning with semantically imagined experience.
\newblock In \emph{Robotics: Science and Systems (RSS)}, 2023.

\bibitem[Yuan et~al.(2025)Yuan, Joshi, Zhu, Su, Zhao, and
  Gao]{yuan2025roboengine}
Chengbo Yuan, Suraj Joshi, Shaoting Zhu, Hang Su, Hang Zhao, and Yang Gao.
\newblock Roboengine: Plug-and-play robot data augmentation with semantic robot
  segmentation and background generation.
\newblock \emph{arXiv preprint arXiv:2503.18738}, 2025.

\bibitem[Zelikman et~al.(2023)Zelikman, Lorch, Mackey, and
  Kalai]{zelikman2023stop}
Eric Zelikman, Eliana Lorch, Lester Mackey, and Adam~Tauman Kalai.
\newblock Self-taught optimizer ({STOP}): Recursively self-improving code
  generation.
\newblock \emph{arXiv preprint arXiv:2310.02304}, 2023.

\bibitem[Zhang et~al.(2025)Zhang, Hu, Lu, Lange, and Clune]{zhang2025dgm}
Jenny Zhang, Shengran Hu, Cong Lu, Robert Lange, and Jeff Clune.
\newblock Darwin g\"odel machine: Open-ended evolution of self-improving
  agents.
\newblock \emph{arXiv preprint arXiv:2505.22954}, 2025.

\bibitem[Zhang et~al.(2026{\natexlab{a}})Zhang, Huang, Deng, Yu, Chen, Ling,
  Xie, and Zhou]{zhang2026mhla}
Kewei Zhang, Ye~Huang, Yufan Deng, Jincheng Yu, Junsong Chen, Huan Ling, Enze
  Xie, and Daquan Zhou.
\newblock {MHLA}: Restoring expressivity of linear attention via token-level
  multi-head.
\newblock In \emph{International Conference on Learning Representations
  (ICLR)}, 2026{\natexlab{a}}.

\bibitem[Zhang et~al.(2026{\natexlab{b}})Zhang, Wang, Gao, Wu, Cao, Han,
  et~al.]{zhang2026fastddrive}
Kewei Zhang, Jin Wang, Sensen Gao, Chengyue Wu, Yulong Cao, Songyang Han,
  et~al.
\newblock {Fast-dDrive}: Efficient block-diffusion {VLM} for autonomous
  driving.
\newblock \emph{arXiv preprint arXiv:2605.23163}, 2026{\natexlab{b}}.

\bibitem[Zhang et~al.(2023)Zhang, Rao, and Agrawala]{zhang2023controlnet}
Lvmin Zhang, Anyi Rao, and Maneesh Agrawala.
\newblock Adding conditional control to text-to-image diffusion models.
\newblock In \emph{IEEE/CVF International Conference on Computer Vision
  (ICCV)}, 2023.

\bibitem[Zhang and Shah(2026)]{zhang2026veritas}
Mingtong Zhang and Dhruv Shah.
\newblock Visual verification enables inference-time steering and autonomous
  policy improvement.
\newblock \emph{arXiv preprint arXiv:2606.18247}, 2026.

\bibitem[Zheng et~al.(2023)Zheng, Chiang, Sheng, Zhuang, Wu, Zhuang, Lin, Li,
  Li, Xing, Zhang, Gonzalez, and Stoica]{zheng2023judging}
Lianmin Zheng, Wei-Lin Chiang, Ying Sheng, Siyuan Zhuang, Zhanghao Wu, Yonghao
  Zhuang, Zi~Lin, Zhuohan Li, Dacheng Li, Eric~P. Xing, Hao Zhang, Joseph~E.
  Gonzalez, and Ion Stoica.
\newblock Judging {LLM}-as-a-judge with {MT}-bench and chatbot arena.
\newblock In \emph{Advances in Neural Information Processing Systems (NeurIPS)
  Datasets and Benchmarks Track}, 2023.

\end{thebibliography}

\newpage
\appendix
\section{Execution harness and agent runtime}
\label{sec:app-harness}
For every attempt, the execution harness checks that the produced document has the stage's IR type, names its producer,
lists exactly its accepted inputs as predecessors, and passes schema validation.
It also checks that the receipt binds the stored digest and comes from a
verifier whose identity differs from the producer's. Only a \textsc{pass} with
no failed check advances; \textsc{reject} and \textsc{abstain} stay in the
history and count against the budget. A rejecting receipt may route the
failure to the earliest responsible ancestor. This opens a new epoch that
invalidates that stage and its dependents while keeping all prior attempts, and
a plan-level reroute budget bounds the number of rewinds. Runs are crash safe:
attempts are written ahead, state is re-verified from stored artifacts on
resume rather than trusted, and a per-run lock prevents duplicate execution.
Paid provider jobs are resumed by their recorded IDs, never resubmitted.

The stages are executed by agents in separate roles (Table~\ref{tab:roles}).
A context compiler expands each role's instructions and skills from the pinned
capability layer into explicit system and user messages and records the hash
of every loaded file and message. Activation therefore does not depend on a host
discovering hidden directories, and every session's instructions can be
reproduced. The orchestrator keeps a resumable task state with a phase, next
action, open questions, and external job IDs. Its instructions allow it to mark a task
\texttt{needs-input} only for intent, device facts, or access that cannot be
derived, and the runtime accepts the mark only with concrete open questions. A missing
adapter or prompt never qualifies, because generating those is its job. \texttt{complete} is accepted only when the delivery verifier
passes, never on the agent's word.

\section{Role instructions}
\label{sec:app-roles}
Table~\ref{tab:roles} lists the agent roles, what each produces, and what each is
not allowed to do.

\begin{table*}[t]
\centering
\small
\resizebox{1.0\textwidth}{!}{%
\begin{tabular}{p{0.19\linewidth}p{0.37\linewidth}p{0.34\linewidth}}
\toprule
Role & Produces & Cannot \\
\midrule
Orchestrator & grounding, task artifacts, render contract, batches, delivery & sign its own reviews or delivery \\
Task designer & task contract, stages, randomization axes & author the expert \\
Task verifier & frozen judge spec and negative controls & be edited by the expert author \\
Expert author & expert controller inside the harness & read the judge internals or holdout seeds, launch rollouts directly \\
Render reviewer / verifier auditor & decoded-media review of motion, topology, timing, and appearance & reuse producer summaries as evidence \\
Pipeline improver & candidate changes to the capability layer & evaluate or activate its own changes \\
\bottomrule
\end{tabular}%
}
\caption{The seven role instructions (render reviewer and verifier auditor
share a row) are loaded explicitly from the capability layer; reviews requested by
the agent runtime use the task-verifier instruction. For tasks run by the agent
runtime, independence is enforced by separate executions and content-bound review
records, not by different names in a JSON field. Reviewers run with the author's
file permissions, so an edit to a reviewed artifact is detected rather than prevented.}
\label{tab:roles}
\end{table*}

\section{Authority factorization}
\label{sec:app-authority}
Each fact in a delivered episode has exactly one source that may certify it
(Table~\ref{tab:authority}). The real episode is authoritative only for what it shows
or measures, so physical parameters that it does not determine remain labelled as
fitted, assumed, or unknown, together with their uncertainty. The simulator becomes
authoritative for actions and states only after it has been qualified for the skills
of the task, whereas the renderer may change only the appearance that the data mode
permits, and a real reference image supplies appearance but never positions, layout,
motion, or additional objects. Verifiers and reviewers, finally, may admit or reject
an episode but cannot modify it. Because no agent role or program both proposes and
certifies the same fact, an error made by one component must be caught by another
before it reaches the dataset.

\begin{table}[t]
\centering
\small
\setlength{\tabcolsep}{3.5pt}
\renewcommand{\arraystretch}{1.15}
\resizebox{1.0\coltabwidth}{!}{%
\begin{tabular}{lcccccc}
\toprule
 & \shortstack{Cameras} & \shortstack{Appearance} & \shortstack{Object} & \shortstack{Actions} & \shortstack{States} & \shortstack{Admission} \\
\midrule
Coding agent         & \auProp & \auProp & \auProp & \auProp & \auProp & \auNo   \\
Calibration episode  & \auOwn  & \auOwn  & \auProp & \auNone & \auNone & \auNone \\
Qualified simulator  & \auNone & \auNo   & \auOwn  & \auOwn  & \auOwn  & \auNone \\
Renderer             & \auNone & \auProp & \auNo   & \auNo   & \auNo   & \auNone \\
Real reference image & \auNone & \auOwn  & \auNo   & \auNo   & \auNo   & \auNone \\
Verifiers, reviewers & \auNone & \auNone & \auNone & \auNone & \auNone & \auOwn  \\
\bottomrule
\end{tabular}%
}
\caption{Who may certify what: \auOwn\ certifies, \auProp\ may propose only, \auNo\ may
not change or certify, --- not involved. No component certifies what it produces, and
reviewers may admit or reject an episode but cannot relax the task predicates.}
\label{tab:authority}
\end{table}

\section{Cost model}
\label{sec:app-cost}
The cost model behind Figure~\ref{fig:cost} and Table~\ref{tab:costmodel} counts only
published or measured figures for the farm; its teleoperation comparisons are given at the end of this section.

For the farm, we render at 640$\times$480 with our H3 deployment, which takes
37--53\,s per 192-frame chunk on one B200 at \$3 per GPU-hour, near the lowest published B200 rental
quote (\$2.80 per GPU-hour, October 2026), plus 20\% for worker start-up, recompiles, and failed
jobs as observed in our runs. In these runs, every chunk was screened by a
vision-language model and re-rendered if flagged (1.1--1.8 renders per chunk;
Section~\ref{sec:render}), and every episode is reviewed with Claude Haiku~5.5. The farm's
one-time and contract costs are the corresponding rows of Table~\ref{tab:cost}. Its marginal cost $m$ per admitted
episode divides the rendering, screening, and review cost of an attempt by the admission rate and adds the rollouts
and audit of Table~\ref{tab:cost} per simulated episode, measured on 100 episodes and assumed to scale linearly;
a rejected render is attempted again from the same simulated episode, so this term is not divided by the admission rate.
For $N$ episodes, the farm costs $C_{\mathrm{adapt}}+C_{\mathrm{contract}}+Nm$, or
$\bar c(N)=(C_{\mathrm{adapt}}+C_{\mathrm{contract}})/N+m$ per
episode; the calibration episode of the request is not counted.
New appearance changes only the render contract's text, so $m$ and $\bar c$ are the same
in the original scene and under new appearance.
Seedance's published price corresponds to about \$1 per second of our three-view video.

We also measured the one-time agent cost with other agents and harnesses on three tasks
(Figure~\ref{fig:cost}(b), Table~\ref{tab:agents}). The scope is that of the one-time row of Table~\ref{tab:cost}: from the
calibration episode to five independently admitted simulated episodes, without rendering, at standard API prices,
excluding the orchestrating session and GPU time. The first four rows of Table~\ref{tab:agents} share one protocol: the same
pinned source, one fresh session per phase, the same review and admission rules, and a \$100 cap per task; SoL-Pi~\citep{liu2026solpi} is an
extension of the Pi coding agent.\footnote{\url{https://github.com/NVlabs/SoL-Pi}} These are single runs per task and agent.
The last two rows are earlier runs under a different protocol: the first is our earlier Claude Code run, whose fold towel cost remains in Table~\ref{tab:cost}, and the Codex run includes a stalled first protocol round.

The differences mostly follow token prices: GPT-6.1 Sol made more requests than GPT-6 Astra, at about a fifth of the price.
On identical briefs for the first phase, SoL-Pi cost 5.9\% less than Codex with the same model, so most of the gap between
the two GPT-6 Astra rows comes from the protocol. The runs also did not need the same help or reach the same quality. With
GPT-6.1 Sol, we sent follow-up instructions after producer sessions stopped with no passing source on stack cups and beat drum. On
beat drum, four of its five admitted episodes exceed the robot's 3\,rad/s joint speed limit in a check we ran afterwards,
which its review did not test. A reviewer rejected the first five beat drum episodes with Claude Opus~5.5 for the same defect,
and that row includes the repair and a second batch. The Claude Fable~5.1 row includes \$13.57 of idle session wake-ups that did no
work (\$61.48 on average without them).

\begin{table}[t]
\centering
\footnotesize
\setlength{\tabcolsep}{3pt}
\resizebox{\coltabwidth}{!}{%
\begin{tabular}{lrrrrrr}
\toprule
Agent and harness & Price & Stack cups & Weigh apple & Beat drum & Mean & Attempts \\
\midrule
GPT-6.1 Sol + SoL-Pi & 2/10 & 15.46 & 8.00 & 11.54 & 11.67 & 5, 5, 18 \\
Claude Opus 5.5 + SoL-Pi & 4/20 & 18.22 & 10.33 & 19.25 & 15.93 & 5, 6, 10 \\
GPT-6 Astra + SoL-Pi & 10/50 & 60.97 & 43.19 & 63.44 & 55.87 & 17, 6, 12 \\
Claude Fable 5.1 + Claude Code & 10/50 & 70.60 & 72.46 & 54.95 & 66.01 & 8, 5, 5 \\
\midrule
Claude Opus 5.5 + Claude Code & -- & 30.72 & 28.23 & 26.66 & 28.54 & -- \\
GPT-6 Astra + Codex & 10/50 & 151.42 & 112.16 & 107.72 & 123.77 & -- \\
\bottomrule
\end{tabular}%
}
\caption{One-time agent cost by agent and harness, in USD at standard API pricing, to five independently admitted
simulated episodes. Price: input/output per million tokens. Attempts: capture attempts for the five admitted episodes,
in column order, every attempt counted.}
\label{tab:agents}
\end{table}

For teleoperation in North America, we have no verifiable quote for a custom collection, so we count only labour, in an
estimate made with Claude Opus 5.5 from a web survey and our own log. Our own 100-episode collections of the five tasks took 36--65\,s of operator time per
episode, including resets and pauses, for 11--28\,s of data. We assume an operator wage of \$40 per hour (\$25--50; public
job postings list about \$22--55), 25\% of episodes discarded (20--30\%), supervision and quality review at 45\% of the
operator's cost (40--50\%), and post-processing of \$0.30 per usable episode (up to \$1), the last three from a
vendor's breakdown~\citep{rcsv2026cost}. This gives \$217--358 per usable hour of data across the tasks, \$286 on
average, which we round to \$290, and \$190--520 under the low and high assumptions. It excludes robots, venue, one-time setup,
and operator training. We convert it to a cost per episode at each task's farm episode length, so that both sides
of the comparison cover the same amount of data.

For China, we use the quote of about RMB 800 per hour (\$117), which we read as per usable hour.
It is a full-service price, so the two comparisons are not like for like.

\section{Image statistics}
\label{sec:app-kid}
Section~\ref{sec:res-views} compares the sources by their kernel distance (KID) to real camera frames,
and Table~\ref{tab:kid} gives its value on every task. Each source renders the same simulated episode of the
task, and 150 evenly spaced head-view frames per source are embedded with DINOv2-base and compared with the
real camera frames of the task by the unbiased polynomial-kernel KID. The averages in
Section~\ref{sec:res-views} (0.83 for H3, 0.88 for Seedance, 1.93 for the simulator, and 2.50 with
domain randomization) use stack cups, weigh apple, and pull drawer, the three tasks on which all four sources
render the same episode, since Seedance has no beat-drum render of that episode and the fold-towel H3 render
is of a different towel episode. Farm episodes are closer to the real frames than the simulator on every
task, and domain randomization moves the images further away, because random textures and lights do not
resemble the real scene. As a reference for indistinguishable distributions, the first row of the table
splits the real frames into two halves; the unbiased estimate can be slightly negative.

\begin{table}[t]
\centering
\footnotesize
\setlength{\tabcolsep}{3pt}
\resizebox{1.0\coltabwidth}{!}{%
\begin{tabular}{lccccc}
\toprule
Source & Stack cups & Weigh apple & Beat drum & Pull drawer & Fold towel \\
\midrule
Real (split half) & -0.02 & -0.16 & -0.01 & -0.00 & -0.03 \\
\midrule
Simulator & 2.19 & 1.95 & 3.74 & 1.63 & 1.41 \\
Sim + DR & 2.73 & 2.69 & 4.27 & 2.07 & 1.58 \\
\ours (Seedance) & 0.69 & 1.21 & 0.38 & 0.73 & 0.66 \\
\ours (H3) & 0.93 & 1.03 & 0.27 & 0.53 & 0.76 \\
\bottomrule
\end{tabular}%
}
\caption{KID ($\times 1000$, lower is closer to real) of head-view frames against real frames, per task.
Real frames per task: stack cups 225, weigh apple 32, beat drum 114, pull drawer 1277, fold towel 201. The first row splits the real frames into two halves.}
\label{tab:kid}
\end{table}

\section{Extended related work}
\label{sec:app-related}
This section expands Section~\ref{sec:related-work}; Table~\ref{tab:related} compares
the closest data-generation pipelines.

\paragraph{Real-world data collection.}
Large teleoperated datasets have driven the recent progress of vision-language-action
models, which shows the value of diverse physical
experience~\citep{kim2024openvla,black2024pi0,black2025pi05,fu2026stablevla,zhang2026fastddrive,ma2026humanscale,gong2026vigar}. Subsequent work lowers the
cost of collection, either with hand-held interfaces that record demonstrations without
a robot~\citep{chi2024umi} or with foundation-model orchestrators that propose tasks and
coordinate fleets of robots~\citep{ahn2024autort}. Each episode, however, still
requires a person, a device, and a physical scene, so the cost grows with the number of
episodes and starts over for every new task. \ours instead requires a single
calibration episode of the scene and produces every episode of the dataset without
robot or operator time.

\paragraph{Demonstration generation from a few demonstrations.}
A second line of work multiplies a few demonstrations into many. MimicGen and its
successors adapt object-centric segments of a few demonstrations to new object
configurations in simulation~\citep{mandlekar2023mimicgen,
garrett2024skillgen,jiang2025dexmimicgen}, and DemoGen edits a single real
demonstration in point-cloud space~\citep{xue2025demogen}. These methods are
efficient, but they assume a simulation task that people have already built or
restrict observations to point clouds. \ours, in contrast, reuses no recorded action:
it builds the simulation task from the request and the calibration episode, generates
new trajectories with an expert written by the agent, and outputs photographic
multi-view video that an RGB policy consumes directly.

\paragraph{Language models that write simulation tasks.}
Language models are increasingly used to automate the construction of robot
learning tasks. They write policies as code~\citep{liang2023codeaspolicies,si2026fewertokens}, design
rewards~\citep{ma2023eureka}, and generate tasks together with their simulation
supervision~\citep{wang2023gensim,wang2023robogen}, and RoboTwin~2.0 closes a
code-repair loop between a multimodal model and the simulator to generate bimanual
data at scale~\citep{chen2025robotwin2}. These tasks, however, are specified by text
and asset libraries rather than by a real scene, their observations are simulator
renderings, and the model that writes a task often also writes the check that grades
it. In \ours, the agent grounds the task in a recording of the real scene and writes
the expert against a judge it cannot see, while independent reviewers decide which
episodes are admitted and the appearance comes from the real scene.

\paragraph{Real-to-sim-to-real and digital twins.}
Real-to-sim-to-real methods narrow the visual gap by reconstructing the target
scene in simulation. RialTo trains policies in reconstructed digital
twins~\citep{torne2024rialto}, URDFormer predicts articulated simulation models from
images~\citep{chen2024urdformer}, and digital-cousin systems trade exact replicas
for diverse, affordance-preserving scenes~\citep{dai2024digitalcousins,
ranawaka2026simfoundry}. Gaussian splatting further makes reconstructed scenes
photorealistic: SplatSim replaces meshes with splats inside a physics
simulator~\citep{qureshi2025splatsim}, RoboGSim and Re$^3$Sim combine reconstruction
with a physics engine to synthesize
demonstrations~\citep{li2024robogsim,han2025re3sim}, Real2Render2Real renders object
scans along trajectories retargeted from a human video~\citep{yu2025real2render2real},
and Video2Sim2Real refines object-centric anchors from video in
simulation~\citep{han2026video2sim2real}. In all of these pipelines, however,
appearance is only as good as the reconstructed assets: every object and scene must be
scanned or modelled, deformable objects and state-dependent appearance such as cloth or
a lit display remain hard to represent, and a change of scene requires a new capture.
\ours keeps the simulator for actions and states but takes appearance from a video
model conditioned on real images, so it needs no per-object scan, renders cloth and
displays as they look, and changes surfaces or lighting through text.

\paragraph{Generative augmentation and video models.}
Generative models offer another route to realistic observations~\citep{zhang2026mhla,huang2026dcsae,chen2026mcsparse,liu2026fvattn}. Image-level
augmentation inpaints new objects and backgrounds into robot
data~\citep{yu2023rosie,chen2023genaug,yuan2025roboengine}, but it edits frames
independently and cannot add new motion. Video models lift this restriction in
different ways: DreamGen generates new robot videos and recovers pseudo-actions with an
inverse-dynamics or latent-action model~\citep{jang2025dreamgen}, Cosmos-Transfer and
RoboTransfer generate video under structural controls such as edges and
depth~\citep{nvidia2025cosmostransfer,liu2025robotransfer}, and EmbodieDreamer and CRAFT
turn simulator renderings into realistic video while the simulator supplies the
actions~\citep{wang2025embodiedreamer,chen2026craft}. Robot-video benchmarks
nevertheless show that structure, physics, task adherence, and action completion fail
independently in such videos~\citep{deng2026rbench}, so realism alone does not make
the data usable. \ours shares the idea of rendering simulated trajectories and adds
three components: an agent that builds the simulation task from one calibration
episode, conditioning on that episode's own images so that the output matches the
deployment scene, and independent admission of every long multi-view episode against
its simulated state.

\paragraph{Agentic software engineering.}
Coding agents have become capable of sustained software engineering~\citep{wei2025ghidorah}. With
suitable tool interfaces they resolve real repository issues~\citep{yang2024sweagent},
and self-improving agent systems retain program versions that score better on
external benchmarks~\citep{zelikman2023stop,hu2024adas,zhang2025dgm}. Language models
nevertheless remain unreliable judges of their own output~\citep{huang2024selfcorrect}.
\ours therefore uses such an agent as the task compiler of a data farm but separates
its work from the certification of that work.

\end{document}